\documentclass[]{article}
\usepackage{arxiv}

\usepackage[utf8]{inputenc} 
\usepackage[T1]{fontenc}    
\usepackage{bbding}

\usepackage{tikz}
\usepackage{siunitx}
\usetikzlibrary{fadings}
\usetikzlibrary{patterns}
\usetikzlibrary{shadows.blur}
\usetikzlibrary{shapes}
\usetikzlibrary{backgrounds}
\usetikzlibrary{automata, positioning, arrows, calc}

\usepackage{graphicx}
\usepackage[inline]{enumitem}
\graphicspath{{pdf/}{photo/}{plots/}}
\DeclareGraphicsExtensions{.pdf,.jpeg,.jpg,.png}

\usepackage[hidelinks=true,bookmarks=false]{hyperref}
\usepackage{amsmath,amssymb,amsfonts}
\usepackage{booktabs}
\usepackage{algpseudocode}
\usepackage{algorithm}
\usepackage{wasysym}
\usepackage[export]{adjustbox}
\usepackage[capitalise]{cleveref}
\usepackage{circledsteps}

\newcommand{\V}[1]{\bar{\boldsymbol{\mathbf{\MakeLowercase{#1}}}}}

\newcommand{\MARKERCIRCLE}[1]{\Circled{#1}}
\pgfkeys{/csteps/fill color=white}

\DeclareSIUnit\gramforce{gf}

\title{TRIGGER: A Lightweight Universal Jamming Gripper for Aerial Grasping}

\author{ \href{https://orcid.org/0000-0001-9220-6679}{\includegraphics[scale=0.06]{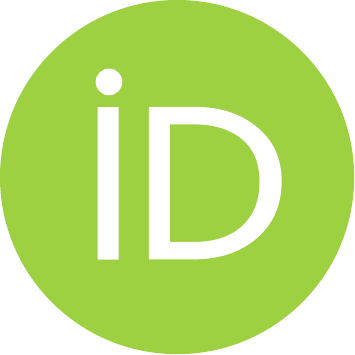}\hspace{1mm}Paul Kremer}\thanks{University of Luxembourg,
    Faculty of Science, Technology and Medicine (FSTM),
	2, Av. de l'Université, L-4365 Esch-sur-Alzette, Luxembourg} \hspace{1mm} \Envelope \\
	\texttt{p.kremer@uni.lu}\\
	\And
	\href{https://orcid.org/0000-0001-9429-7164}{\includegraphics[scale=0.06]{orcid.pdf}\hspace{1mm}Hamed Rahimi Nohooji}\thanks{Interdisciplinary Center for Security, Reliability and Trust (SnT),
	University of Luxembourg,
	29, Av. J.F. Kennedy, L-1855 Luxembourg} \\
    \texttt{hamed.rahimi@uni.lu}\\
 	\And
	\href{https://orcid.org/0000-0001-5018-0925}{\includegraphics[scale=0.06]{orcid.pdf}\hspace{1mm}Jose Luis Sanchez-Lopez}\footnotemark[2]\\
    \texttt{joseluis.sanchezlopez@uni.lu} \\
  	\And
	\href{https://orcid.org/0000-0002-9600-8386}{\includegraphics[scale=0.06]{orcid.pdf}\hspace{1mm}Holger Voos}\footnotemark[1] \hspace{0.5mm} \footnotemark[2]\\
    \texttt{holger.voos@uni.lu} \\
}

\begin{document}
\maketitle

\begin{abstract}
This work introduces TRIGGER, the first ligh\textbf{T}weight unive\textbf{R}sal jamm\textbf{I}n\textbf{G} \textbf{G}ripper for a\textbf{E}rial g\textbf{R}asping.
TRIGGER is an omnidirectional, landing-capable aerial grasping system with resilience and robustness to collisions and inherent passive compliance.
In particular, this work presents the design, fabrication, and experimental validation of a novel, intelligent, modular, universal jamming gripper specifically designed for aerial applications. 
Leveraging recent developments in particle jamming and soft granular materials, TRIGGER produces \SI{15}{\newton} of holding force with only a relatively small activation force of \SI{2.5}{\newton}.
Experiments show the relationship between fill ratio and activation force and reveal that adding an additive to the membrane's silicone mixture improves the holding force by up to \SI{52}{\percent}.
The concept is validated by mounting TRIGGER onto a multicopter performing a pick-and-release task under laboratory conditions.
Based on the experimental data, a model for robotic simulators is introduced to facilitate future controller developments. 


\end{abstract}



\section*{License}
For the purpose of Open Access, the author has applied a CC-BY-4.0 public copyright license to any Author Accepted Manuscript version arising from this submission.

\section{Introduction}
\label{sec:introduction}

During the last decade, Unmanned Aerial Vehicles (UAVs) attracted considerable research interest from the robotics community.
In 2010 seminal work on fully functional UAVs for plant inspection for UK onshore oil refineries \cite{Shukla2016} was developed.
Only a few years later, UAV technology had matured to a point where it could be utilized in several applications such as remote sensing and monitoring of forest fires \cite{Casbeer2006}, agricultural products \cite{Faical2014}, border monitoring \cite{Beard2006}, search and rescue \cite{Almurib2011}, plant assets inspection \cite{Connolly2014}, and transmission line inspection \cite{Li2013a}.
Despite the huge success UAVs had in sensing and monitoring applications, they fell short when it came to tasks that required physical interaction with their surroundings.
Consequently, the field of aerial manipulation emerged, addressing those deficits by equipping UAVs with robotic manipulators \cite{BonyanKhamseh2018a} and claw-like grippers \cite{Popek2018}, \cite{Kruse2018}.
In the last few years, aerial manipulation has considerably evolved, featuring numerous applications, including medical goods delivery \cite{Thiels2015}, infrastructure monitoring and maintenance \cite{Rubiales2021}, \cite{Bodie2019}, autonomous transportation and construction \cite{Loianno2018}, agriculture \cite{Ore2015}, and forestry \cite{Kaslin2018}.


\begin{table*}[t]
\resizebox{1.0\linewidth}{!}{%
\begin{tabular}{llccclm{1.2cm}cm{1.2cm}m{1.2cm}c}
\toprule
Drone Gripper                                & Archetype          & $m$ (kg)     & $F_h$ (N)  & $F_h/m$ & Actuation Method        & Sensors            & Compliant   & Omni-directional          & Landing capable           & Year \\ \midrule
Self-sealing suction cup \cite{Kessens2016a} & suction cup        & 0.72          & 12        & 1.7     & pneumatic (pump)        & pressure           & -           & \centering \checkmark     & \centering  -             & 2016 \\
Permanent Magnet Hand \cite{Fiaz2018a}       & magnet             & 0.30          & 25.48     & 8.7     & magnetic                & contact            & -           & \centering \checkmark     & \centering  -             & 2018 \\
Actively Compliant Gripper \cite{Kruse2018}  & arm+claw           & 0.30          & 0.57      & 0.2     & servo (tendon)          & /                  & \checkmark  & \centering -              & \centering -              & 2018 \\
Soft Grasper \cite{Mishra2018}               & hand               & 0.58          & 10-20     & 3.5     & pneumatic (cartridge)   & /                  & \checkmark  & \centering -              & \centering -              & 2018 \\
Small Sleeved Gripper \cite{Miron2018}       & hand               & 0.38          & 52        & 14.0    & pneumatic (cartridge)   & /                  & \checkmark  & \centering -              & \centering \checkmark     & 2018 \\
Ultra-fast Robot Hand \cite{McLaren2019}     & hand               & 0.55          & 51        & 9.5     & servo (tendon)          & proximity          & \checkmark  & \centering -              & \centering -              & 2019 \\
Mechatronic Jaw Gripper \cite{Lieret2020a}   & claw               & N/A           & 2         & N/A     & servo (direct)          & aperture           & -           & \centering -              & \centering -              & 2020 \\
Soft-Tentacle Gripper \cite{Rubiales2021}    & hand               & 0.008         & 2.12      & 27.0    & passive                 & /                  & -           & \centering -              & \centering \checkmark     & 2021 \\
Micro Bistable Gripper \cite{Zhang2021b}     & claw               & 0.008         & 2.12      & 27.0    & passive                 & /                  & -           & \centering -              & \centering \checkmark     & 2021 \\
Hybrid Suction Cup \cite{Tsukagoshi2021}     & suction cup        & 0.050         & 80        & 163.1   & pneumatic (pump)        & pressure           & -           & \centering \checkmark     & \centering \checkmark     & 2021 \\
RAPTOR \cite{Appius2022}                     & claw               & N/A           & N/A       & N/A     & servo (direct)          & /                  & \checkmark  & \centering -              & \centering  -             & 2022 \\
HASEL Gripper \cite{Tscholl2022}             & hand               & N/A           & 0.8       & N/A     & hydraulic/electrostatic & /                  & \checkmark  & \centering -              & \centering  -             & 2022 \\\midrule
TRIGGER                                      & UG                 & 0.38          & 15        & 4.0     & pneumatic (pump)        & force, pressure    & \checkmark  & \centering \checkmark     & \centering \checkmark     & 2023 \\\bottomrule
\end{tabular}%
}
\caption{Comparison of robotic grippers developed for drones: archetype; the total mass $m$; maximum holding force $F_h$; $F_h/m$ the holding force to mass ratio; actuation method; integrated sensors; passive compliance; omnidirectional (indifferent to grasping direction); capability to serve as landing gear; year of publication.}
\label{tab:gripper-comparison}
\end{table*}

Several drone-grippers featuring a wide variety of grasping mechanisms were developed over the past years (see \cref{tab:gripper-comparison}), e.g., pneumatic soft fingers \cite{Miron2018}, rigid jaw grippers \cite{Lieret2020a}, passive bi-stable grippers for micro UAVs \cite{Zhang2021b}, suction cups \cite{Kessens2016a}, magnets \cite{Fiaz2018a}, and flexible limb grippers \cite{Ma2013}.
Some of these grippers are passively compliant, e.g., the soft finger-based grippers \cite{Miron2018}, while others are completely rigid, e.g., \cite{Fiaz2018a} and \cite{Lieret2020a}.
In practice, both approaches are viable, and the optimal choice of the gripper is ultimately application dependant \cite{DAvella2020}.
As a rule of thumb, rigid grippers are less versatile since they are made for specific payloads (e.g., box-shape objects \cite{Lieret2020a}, or for specific materials \cite{Fiaz2018a}), heavier due to mechanical joints, but in turn, can provide a more secure grasp for the payloads they are optimized for.
On the other hand, soft grippers can grasp a broad variety of payloads \cite{Miron2018} by passively conforming to the grasped objects.
Their soft structure allows them to soften the impact and thus reduce the contact forces that can potentially destabilize the drone.
Furthermore, soft grippers are generally more tolerant towards position errors which are inevitable due to the ground effect \cite{Pounds2010}.
These features make soft grippers very promising candidates for grasping in complex, unstructured environments \cite{Milana2022}.
There is however no free lunch, and the gain in versatility is often paid for by compromising in other areas, e.g., cycle time.



In \cite{Brown2010}, a very particular type of soft gripper was introduced, namely, the \textit{Universal Jamming Gripper} (UG), which is a pneumatic gripper based on the jamming principle of certain granular materials. 
This UG, being in essence just a bag containing granular material, works via three distinct mechanisms that are simultaneously involved in the grasping process, namely geometric interlocking $F_G$, suction $F_S$ and friction $F_R$, which all come into play once the soft membrane of the gripper is pressed against the payload by a force called \textit{activation force}.
After jamming (hardening) of the granular material \cite{Chakraborty2018}, the resulting \textit{holding force} is then the sum of all components: $F_h = F_G + F_S + F_R$. 
The jamming typically involves creating a vacuum inside the membrane; however, other jamming principles exist, e.g., by magnetic fields \cite{Nishida2016}, or hydraulic fluids \cite{Sakuma2018}.
As discussed in \cite{Amend2012}, UGs have virtually infinite degrees of freedom that do not need to be controlled explicitly, which gives them the characteristic of being able to grip objects of vastly different shapes thanks to the passively compliant membrane.
Given their symmetric shape, they have no preferred in-plane (horizontal) grasping direction and are thus omnidirectional.
UGs tolerate relatively large positional and angular (tilt) errors during the grasp \cite{Brown2010}.
It is shown in \cite{Amend2012} that off-center grasping with a positional error of up to \SI{60}{\percent} of the membrane's radius does not degrade the gripper's grasping capability.
The versatility and the relaxed requirements for positional and angular accuracy and, consequently, less stringent control requirements serve as the main motivation for developing the UG described herein.
The structure of the UG is well suited to double as the drone's landing gear (contrary to the ubiquitous soft finger grippers).
Furthermore, UGs provide rigid-like grasps by the jamming of the granular material.

\begin{figure}[ht!]
    \centering
    \resizebox{0.7\linewidth}{!}{%
    \includegraphics[frame]{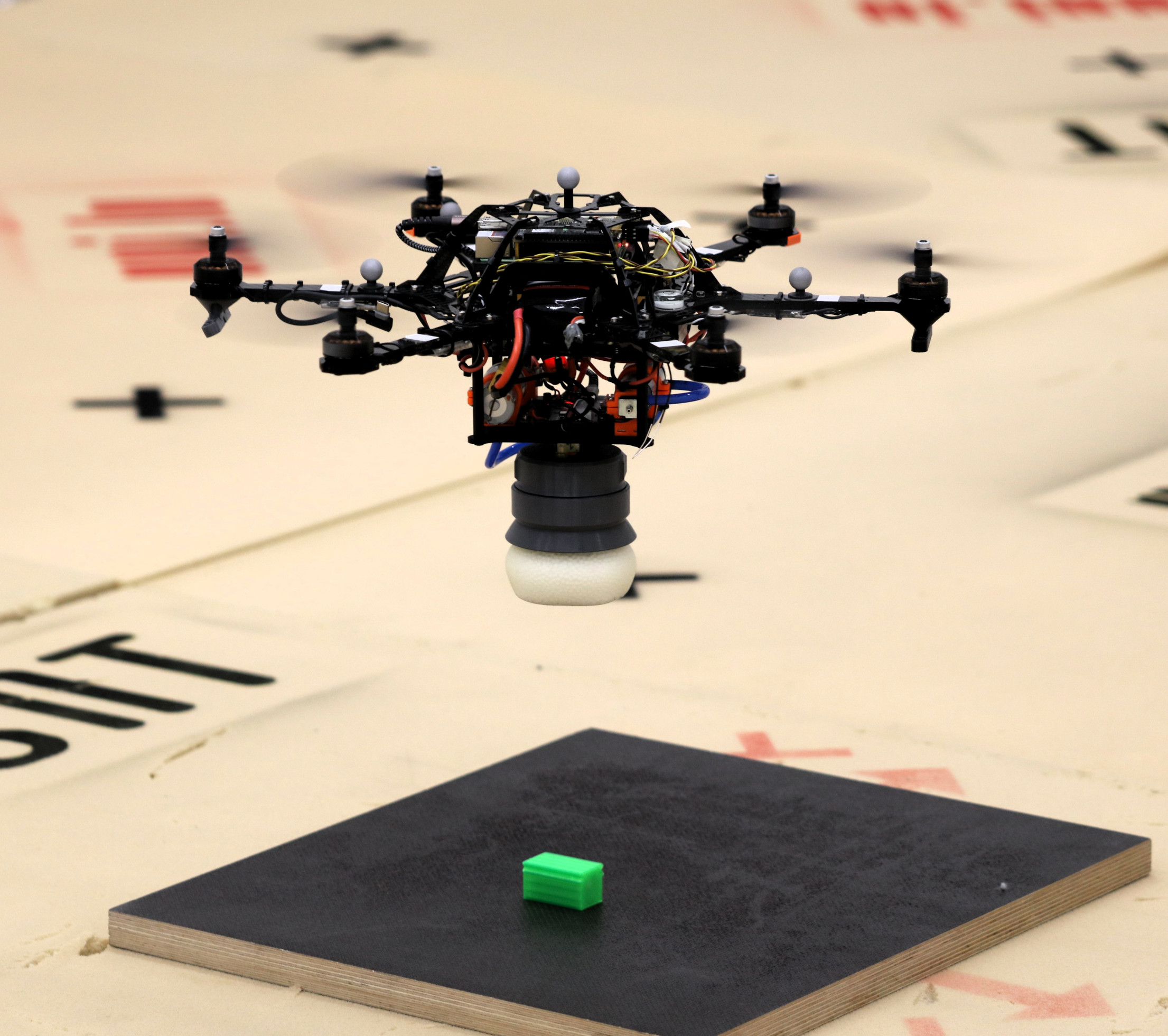}%
    }
    \caption{General aerial grasping concept. TRIGGER is attached to the UAV's cargo bay, similar to a claw-gripper. The UAV's landing gear has been removed as the gripper assures this functionality. The box (green) serves as the dummy payload.}
    \label{fig:concept}
\end{figure}

This work is the first study that adapts the original concept of the \textit{Universal Jamming Gripper} presented in \cite{Brown2010} for aerial manipulation.
By introducing TRIGGER (ligh\textbf{T}weight unive\textbf{R}sal jamm\textbf{I}n\textbf{G} \textbf{G}ripper for a\textbf{E}rial g\textbf{R}asping), we provide a design and implementation of a UG with pneumatic actuation in a small and compact form factor, suitable for small to medium-sized multicopters (see \cref{fig:concept}). 
Our goal is to address some of the open challenges, particularly in grasping scenarios where only very few assumptions can be made about the shape, size, weight, and position of the payload, which paves the way for a variety of practical applications of this type of gripper in the context of aerial grasping. 
This work shows that UGs are inexpensive and easy to build due to their simple construction and offer distinct advantages over other drone grippers.

The main contributions of this work are as follows:
\begin{enumerate}
    \item To the best of the authors' knowledge, this is the first work that conceptualizes, discusses and uses UGs in aerial manipulation.  
    UGs can relax the aerial vehicle’s dependency on the grasping direction and the required positional accuracy.
    Therefore, they address open challenges regarding aerial grasping in complex environments \cite{Milana2022}.

    \item We propose a new design of a UG called TRIGGER, a lightweight and compact gripper for aerial manipulation. 
    The proposed design shares many advantages of available soft grippers in the field, e.g., \cite{Fishman2021}, \cite{Appius2022}, such as resilience and robustness to collisions and the inherent passive compliance, which decouples the UAV from the environment. 
    However, the salient features of the proposed system lie in the intuitiveness of the design, in the simplicity of its omnidirectional grasping mechanism and in its ability to also act as landing gear.
    Our grasping system is modular, energy-efficient, and highly integrated while still being structurally simple and inexpensive to fabricate.
    
    
    \item We provide extensive experimental validation of our design in conjunction with our custom test jig. 
    By analyzing the relation between activation force and fill ratio, we designed TRIGGER to work with a much lower activation force than traditional UGs (e.g., \cite{Kapadia2012}, \cite{GomezPaccapelo2021}), making it suitable for small to medium-sized aerial platforms.
    Furthermore, we show that the holding force can be substantially increased with the help of a silicone additive.
    Using the collected data, we then propose a model of our UG for use in robotic simulators. 
    
    \item Lastly, we show TRIGGER attached to a multicopter, performing a pick and release task, validating the overall concept.
\end{enumerate}

The rest of this work is organized as follows. 
\cref{sec:concept-architecture} introduces the main design challenges and solutions associated with the design and manufacturing of TRIGGER. 
In \cref{sec:experiments}, TRIGGER is characterized followed by the presentation of the experimental results regarding the activation force and the impact of the silicone additive on the holding force.
Based on the experimental data, a model of the developed gripper for robotic simulators is proposed in \cref{sec:digital-twin}.
\cref{sec:aerial-application} showcases TRIGGER in an aerial application.
\cref{sec:discussion} discusses the design and main findings. 
This work is summarized in \cref{sec:conclusion}.

\section{Concept and System Architecture}
\label{sec:concept-architecture}
In this section, we first introduce the general concept of our UG and the associated design challenges in the context of aerial manipulation.
We then take an in-depth look at the electro-mechanical, pneumatic, and software components.

\subsection{Concept}
Multirotor platforms come with many benefits but also with a set of limitations. The most relevant ones are their limited payload capability, the constrained volume for attachments, their underactuated nature, and the challenging dynamics coupling. 
The dynamics coupling is particularly important for aerial systems carrying manipulators \cite{Kremer2022}, 
but it also poses a problem for simpler 'claw' setups where only the grasping element gets in contact with the environment.
Elastic elements inserted in the construction of the grasping device efficiently reduce the dynamic coupling by softening the hard socks associated with typical grasping operations. 
Those elastic elements are inherently present in UGs as represented by their soft membrane.
UGs are thus an ideal fit, provided they can be constructed to fit the size, weight and power envelope of aerial platforms.

Our proof-of-concept aerial platform is a medium-sized (wheelbase of \SI{430}{\milli\meter}), modified \textit{AscTec Firefly} hexacopter with a maximum payload capacity of \SI{1}{\kilo\gram}. 
This airframe conveniently features a cargo bay measuring $\SI{120}{\milli\meter} \times \SI{120}{\milli\meter}$, which is used as the anchor point for TRIGGER.
This particular mounting scheme with the gripper oriented toward the bottom is commonly called a 'claw'.

Compatibility with state-of-the-art autopilots (e.g., Pixhawk) is assured by either directly connecting the gripper to the autopilot via UART or by connecting it to the corresponding companion computer using USB.
For ease of integration, our concept envisions being directly powered by the UAV's main 3S-4S battery, which eliminates carrying an additional battery.
Lightweight construction, modularity, and tight integration of the electronics, the sensors, and the software are the driving concepts of TRIGGER.

To make our work easily reproducible, accessible, and low-cost (below $\$100$, without the manufacturing equipment), we limited our design to widely available and inexpensive manufacturing techniques, where a Fused Deposition Modeling (FDM) printer and a high-power single-stage vacuum pump represent the bulk of the cost. Furthermore, we designed our grasper around customary off-the-shelf parts.

\begin{figure}
    \centering
    \resizebox{0.7\linewidth}{!}{%
    \input{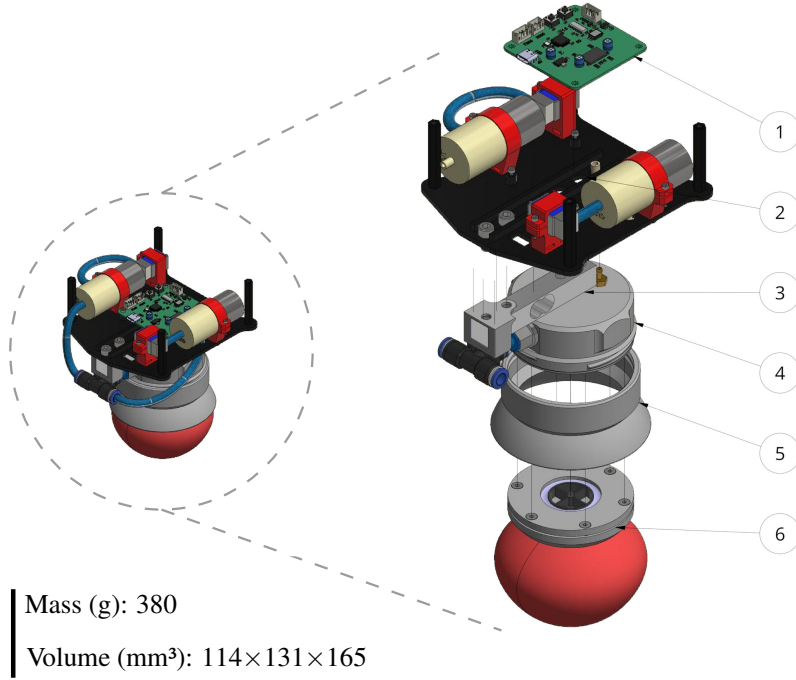}%
    }
    \caption{Complete assembly of the grasping system. \MARKERCIRCLE{1} Custom controller board, \MARKERCIRCLE{2} base assembly with pneumatic system, \MARKERCIRCLE{3} load cell, \MARKERCIRCLE{4} gripper-floor, \MARKERCIRCLE{5} wedge with thread, \MARKERCIRCLE{6} membrane module}
    \label{fig:gripper-asm} 
\end{figure}

The complete grasping system is detailed in \cref{fig:gripper-asm}. 
Its major subsystems are explained hereafter.

\subsection{Pneumatics and Mechanics}
The role of the pneumatic system is twofold: \begin{enumerate*}
    \item to pressurize the membrane and thus allow the contained granular material to flow easily within the free, air-filled volume;
    \item to vaccumize the membrane and consequently jam (i.e., solidify) the granular material
\end{enumerate*}.

UGs can be realized in two distinct topologies, i.e., either as closed-loop or open-loop systems. 
In closed-loop systems, the fluid surrounding the granular material stays contained within the system. An example of such a system is the magnetorheological fluid-based UG shown in \cite{Nishida2016} for the hydraulic UG presented in \cite{Sakuma2018}.
Generally, these systems have the main disadvantage that the fluid has to stay contained within the system (e.g., in tanks that add weight and cost) and that leakage must be considered as a critical failure mode.
On the other hand, open-loop systems exchange their fluid with their environment. 
The operating fluid in that case is thus typically air, resp. water for underwater applications \cite{Licht2016}.
Those systems have the salient advantage that their fluid is abundantly present in their surroundings, which eliminates the storage needs and reduces the severity of leakage, e.g., due to membrane rupture.
Open-loop systems are generally better suited for lightweight construction and require less engineering effort.

\begin{figure}
    \centering
    \resizebox{0.50\linewidth}{!}{%
    \input{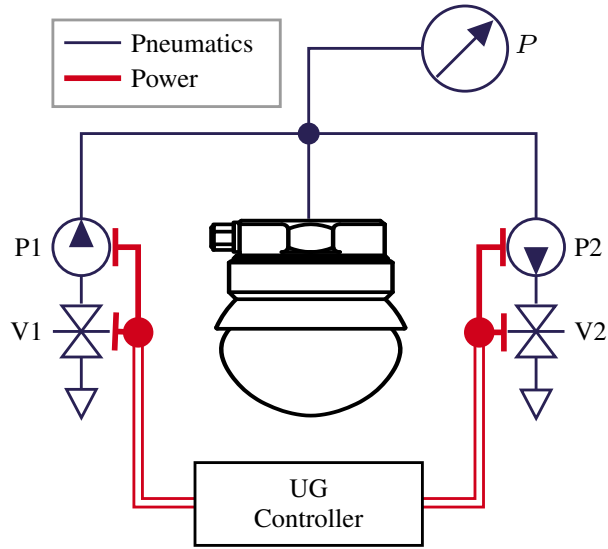}%
    }
    \caption{Pneumatic system. The jamming gripper is fed by two pumps. Pump P1 inflates the balloon and fluidizes its content. Pump P2 creates a vacuum strong enough to jam the particles inside the balloon. The solenoid valves V1, V2 are required to prevent air from leaking through the inactive pumps. All pneumatic components are actuated based on the control inputs from the main controller.}
    \label{fig:pneumatic-arrangement} 
\end{figure}

Therefore, the pneumatic system presented in this paper (\cref{fig:pneumatic-arrangement}) has an open-loop structure and uses air as its operating fluid.
It consists of two small, non-reversible diaphragm pumps (P1, P2) coupled to two pneumatic solenoid 2/1-way valves (V1, V2). 
The air pressure in the system is measured by the Microelectromechanical System (MEMS) pressure sensor $P$.
This particular setup is very low cost and has a favorable mass distribution due to symmetry.
By design, diaphragm pumps act as one-way check-valves, not restricting the airflow in their nominal direction, which therefore requires closing the valve associated with the antagonistic pump such that they can establish a pressure differential.
This particular topology also permits to seal off the system.
The membrane can thus remain pressurized (resp. in a state of vacuum) without powering the pumps, which saves energy.

We utilize two 12V, \SI{7}{\watt}, \texttt{SC3704PM} diaphragm pumps rated for a pressure differential of \SI{46}{\kilo\pascal} at \SI{2}{\liter\per\minute}. The miniature 2/1 air valves are of type \texttt{SCO520FVG}.
Our low-power pneumatic system typically consumes less than \SI{10}{\watt}, contrary to other systems frequently featured in the literature, which are using heavy (more than \SI{1}{\kilo\gram}) stationary, high-power vacuum pumps in the \SI{500}{\watt} range and reaching pressure differentials beyond \SI{80}{\kilo\pascal} \cite{Brown2010}, \cite{Santarossa2021}.
Note that our lower-power system naturally comes with longer cycle times and a lower maximum pressure differential (we measured approximately \SI{28}{\kilo\pascal}), $3\times$ lower than conventional solutions.
However, we will show in \cref{sec:experiments} that this does not adversely affect the performance of the UG.

\begin{figure}
    \centering
    \resizebox{0.6\linewidth}{!}{%
    \includegraphics{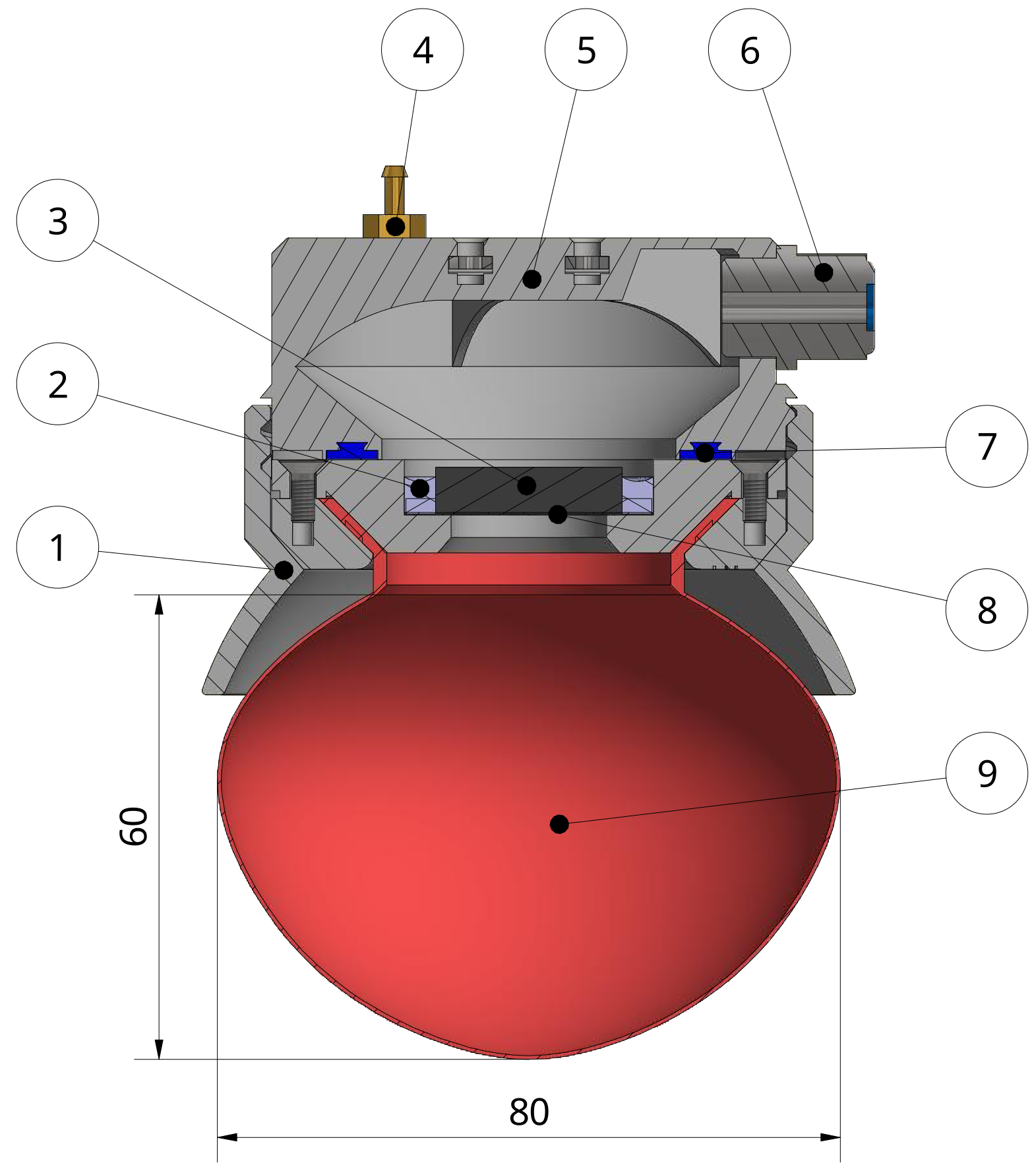}%
    }
    \caption{Section view of the main gripping module (membrane module attached to the gripper-floor). 
    \MARKERCIRCLE{1} wedge with integrated thread,
    \MARKERCIRCLE{2} hot-glue seal,
    \MARKERCIRCLE{3} mechanical filter support,
    \MARKERCIRCLE{4} pressure gauge fitting,
    \MARKERCIRCLE{5} upper shell,
    \MARKERCIRCLE{6} main air inlet/outlet,
    \MARKERCIRCLE{7} cast-in-place silicone gasket,
    \MARKERCIRCLE{8} paper filter,
    \MARKERCIRCLE{9} membrane filled with granular filler material.}
    \label{fig:gripper-balloon-module}
\end{figure}

Concerning the mechanical structure, our modular design approach is shown in \cref{fig:gripper-asm,fig:gripper-balloon-module}. It consists of three larger sub-assemblies, namely \begin{enumerate*}
    \item the base, containing the pumps, valves, and controller board,
    \item the gripper-floor, forming the interface between the pneumatic system and the detachable membrane module,
    \item the membrane module, which firmly holds onto the filled, custom silicone membrane. 
    It contains a paper filter that seals off the filler material from the environment while permitting air to circulate freely.
    A mechanical support structure prevents it from tearing under load.
    The membrane module is firmly pressed against the cast-in-place silicone seal on the gripper-floor by screwing the wedge onto the external printed thread to create an air-tight seal.
\end{enumerate*}

This modular concept has three main advantages: \textit{first}, it enables quick iteration on membrane module designs; \textit{second}, it allows to quickly and effortlessly swap between different membrane modules during the tests; \textit{third}, it enables the platform to be compatible with different types of grippers, given that there are some geometries that cannot be picked up by a UG (e.g., large flat surfaces), which require highly specialized grippers such as vacuum cups.

Our UG is designed to be mounted like a 'claw' on a multirotor; therefore, it does double duty, i.e., it operates as a gripper but also serves as the landing gear. 
As such, it is dimensioned to withstand the total weight and impact of a landing UAV, which comes with several advantages that we discuss in \cref{sec:discussion}.

\subsection{Material Selection}
Our membrane is made from the soft silicone rubber \textit{Trollfactory Type 23}, shore hardness 10 A with \SI{600}{\percent} elongation at break.
The reasons for selecting such a soft rubber are twofold: \textit{first}, it allows us to widen the tolerances on the membrane's thickness as small deviations no longer have a significant impact on the overall stiffness;
\textit{second}, it maximizes the contact area between the membrane and the payload and, therefore, the quality of the grasp is increased. 
Furthermore, this particular silicone can be mixed with a silicon additive called \textit{deadener} (also sometimes referred to as \textit{slacker}), which gives the silicone more human skin-like physical properties.
This further increases the softness of the material and, more importantly, makes it sticky.
The intensity of those effects is controlled by the relative amount of additive added to the mixture.
This specific silicone rubber is very viscous (\SI{14}{\pascal\second}) and thus does not flow easily.
This aspect has to be considered for the mold design and casting process to avoid trapping air inside the mold and thus creating voids in the thin membrane.

For the printed structural parts, PET-G was chosen over PLA for its higher impact resistance and lower density. 
Furthermore, PLA is prone to creep under sustained load. 
The structural parts would not benefit from high-end polymers such as PA6-CF or PEEK as there are no special requirements concerning the stiffness or heat resistance that could motivate such a choice.

Aiming for a lightweight design, we choose EPS as filler material as it has a density of only \SI{17}{\gram\per\liter}, which is by an order of magnitude lower than other commonly used materials such as ground coffee or glass beads (\cref{tab:filler-materials}).
Moreover, the soft EPS particles develop higher holding forces than rigid particles due to the squeezing effect, which is a result of the elasticity of the EPS beads themselves \cite{Santarossa2021}.

\begin{table}
\centering
\begin{tabular}{lll}
\toprule
Material & Density (\si{\gram\per\liter}) & Particle Size (\si{\milli\meter})          \\ \midrule
EPS      & 17                             & 1-4                                        \\
Coffee   & 308                            & 0.2-2                                      \\
Polymer  & 940                            & 0.1-0.2                                    \\
Glass    & 2500                           & 0.2-0.4                                    \\ \bottomrule
\end{tabular}
\caption{Comparison of filler materials. EPS has by far the lowest density.}
\label{tab:filler-materials}
\end{table}

\subsection{Fabrication}
\begin{figure*}
    \centering
    \resizebox{1.0\linewidth}{!}{
    \input{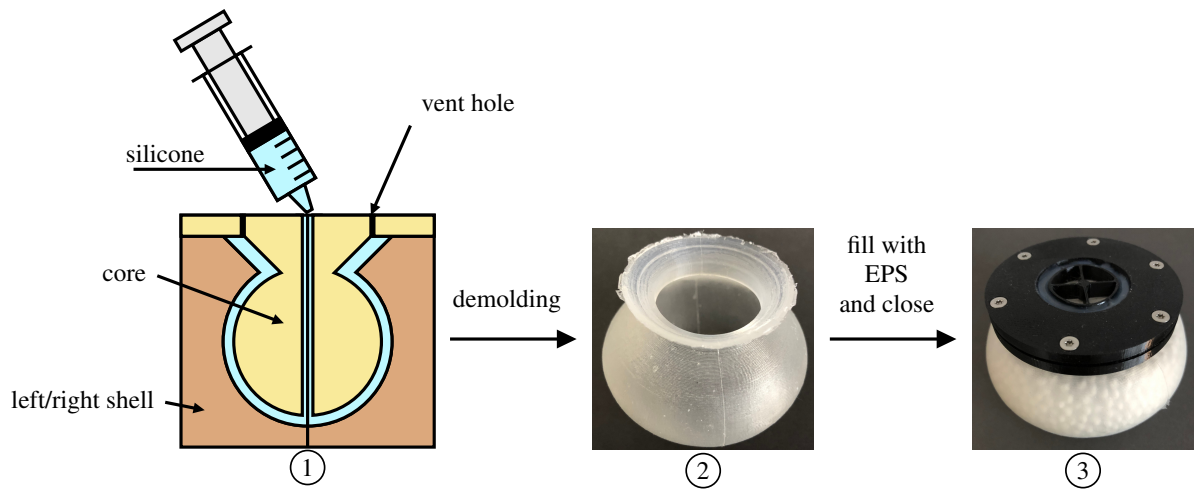}
    }
    \caption{Fabrication process of the membrane. \MARKERCIRCLE{1} The degassed silicone mixture is injected with a syringe through the mold's core. The air inside the mold escapes through the vent holes at the top. Precise alignment between the shells is critical as the membrane is only \SI{0.6}{\milli\meter} thick. \MARKERCIRCLE{2} The obtained membrane after demolding. \MARKERCIRCLE{3} The membrane is filled with EPS and closed with a mechanical assembly that contains the filter. The resulting gripper module is thus fast and easy to swap out in case of damage.}
    \label{fig:mold-casting} 
\end{figure*}

Based on our previous experience with silicone \cite{ZhiliChenHamedRahimiNohooji2016}, we chose a silicone casting process to create the membrane. 
We created a three-part mold (i.e., left and right shell, plus core) printed from PET-G using a common FDM 3D printer. 
Our approach is similar to \cite{Sakuma2018}; however, due to the very thin \SI{0.6}{\milli\meter} membrane and the high viscosity of the silicone rubber, the process had to be adapted. 
More precisely, instead of pouring the silicone into the mold, we inject it directly through the core using a syringe (\cref{fig:mold-casting}). 
This technique enables very thin-walled castings (assuming proper alignment of the shells). 
But, more importantly, it allows the silicone mixture to spread evenly with a fairly low risk of catching air bubbles in the process.
The usual precautions should be taken when working with silicone, such as properly degassing the silicone after mixing.
Our membrane has a nominal diameter of \SI{80}{\milli\meter}, a nominal thickness of \SI{0.6}{\milli\meter}, a height of \SI{60}{\milli\meter}, an encompassing volume of \SI{0.2}{\liter} and a total mass of only \SI{18}{\gram} (without filler).

The structural parts were also fabricated from PET-G using FDM printing. 
The resulting parts have proven to be sufficiently airtight using optimal print settings. 
At the mating point of two structural parts (i.e., \MARKERCIRCLE{5} and \MARKERCIRCLE{9} in \cref{fig:gripper-balloon-module}), a silicone gasket is introduced that assures an air-tight connection between the two parts.

Our filler material consists of a mixture of Expanded Polystyrene (EPS) beads of various sizes ranging from \SI{1}{\milli\meter} to \SI{4}{\milli\meter}. 
Contrary to rigid filler materials such as glass beads, the softness of the particles gives birth to a squeezing effect which is reported to increase the holding force within certain limits \cite{Santarossa2021}. 
Another consideration for the choice of the filler material was the density or, more precisely, the resulting weight of the filled membrane. 
EPS beads have a very low density of approx. \SI{17}{\gram\per\liter} and thus do not add much mass to the system. 
Other materials such as ground coffee with a density of \SI{308}{\gram\per\liter} or glass beads with \SI{2500}{\gram\per\liter} result in significant extra weight.
We added \SI{2.2}{\gram} filler material (\SI{0.13}{\liter}) to the membrane corresponding to a fill ratio of \SI{66}{\percent}.

\begin{figure}
    \centering
    \resizebox{0.5\linewidth}{!}{%
    \includegraphics[]{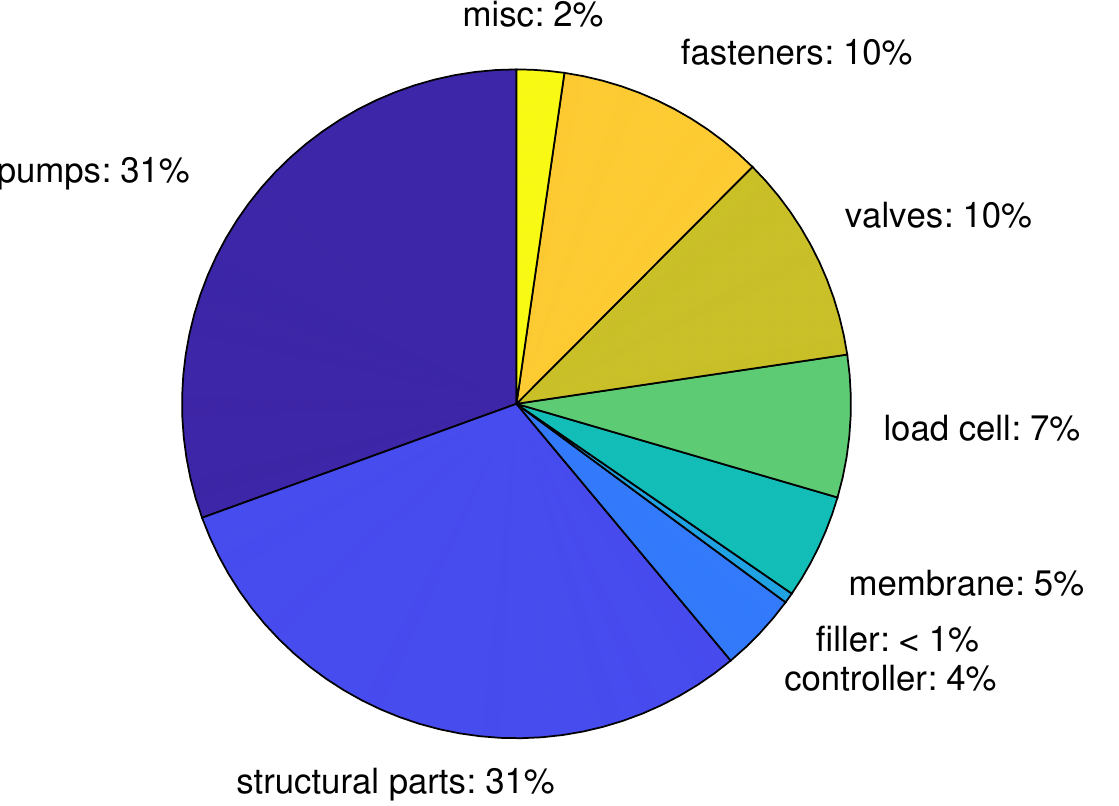}%
    }
    \caption{Mass breakdown of the \SI{380}{\gram}. The pumps and the structural plastic parts form the bulk of the mass.}
    \label{fig:mass-distribution} 
\end{figure}

The total mass of the assembly (\SI{380}{\gram}) is distributed among the different components as shown in \cref{fig:mass-distribution}. 
The pneumatic system represents the bulk of the mass (\SI{160}{\gram}), followed by the structural plastic parts (\SI{115}{\gram}) and the fasteners (\SI{35}{\gram}), fittings and tubing (less than \SI{8}{\gram}).
The mass added by the filler material (\SI{2.2}{\gram}) is, however, negligible.

\subsection{Electronics and Firmware}

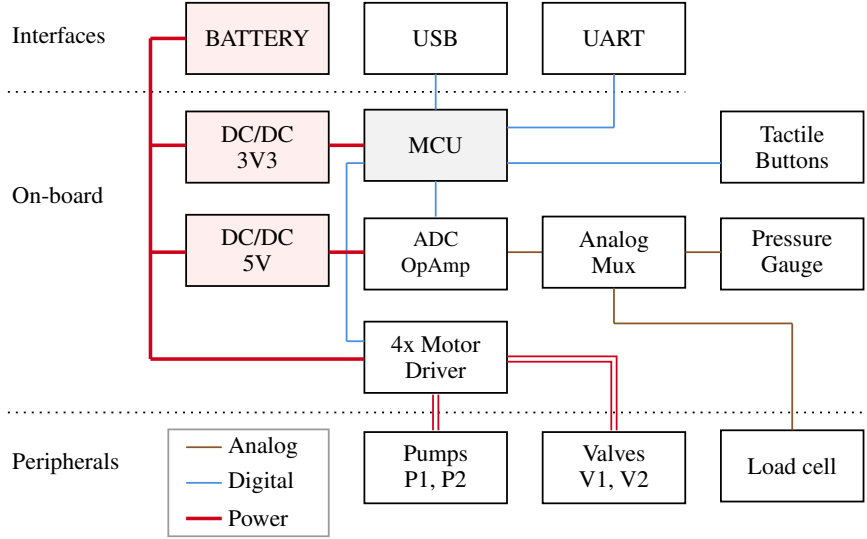
\begin{figure}
    \centering
    \resizebox{0.7\linewidth}{!}{%
    \tikzset{every picture/.style={line width=0.75pt}} 

\begin{tikzpicture}[x=0.75pt,y=0.75pt,yscale=-1,xscale=1]

\draw  [fill={rgb, 255:red, 242; green, 242; blue, 242 }  ,fill opacity=1 ] (210,100) -- (290,100) -- (290,140) -- (210,140) -- cycle ;

\draw   (210,219) -- (290,219) -- (290,259) -- (210,259) -- cycle ;

\draw   (410,280) -- (490,280) -- (490,320) -- (410,320) -- cycle ;

\draw   (410,161) -- (490,161) -- (490,201) -- (410,201) -- cycle ;

\draw   (210,281) -- (290,281) -- (290,321) -- (210,321) -- cycle ;

\draw   (310,281) -- (390,281) -- (390,321) -- (310,321) -- cycle ;

\draw   (210,40) -- (290,40) -- (290,80) -- (210,80) -- cycle ;

\draw   (310,40) -- (390,40) -- (390,80) -- (310,80) -- cycle ;

\draw  [fill={rgb, 255:red, 255; green, 238; blue, 238 }  ,fill opacity=1 ] (110,159) -- (190,159) -- (190,199) -- (110,199) -- cycle ;

\draw  [fill={rgb, 255:red, 255; green, 238; blue, 238 }  ,fill opacity=1 ] (110,40) -- (190,40) -- (190,80) -- (110,80) -- cycle ;

\draw  [fill={rgb, 255:red, 255; green, 238; blue, 238 }  ,fill opacity=1 ] (110,101) -- (190,101) -- (190,141) -- (110,141) -- cycle ;

\draw  [dash pattern={on 0.84pt off 2.51pt}]  (10,90) -- (390,90) ;
\draw  [dash pattern={on 0.84pt off 2.51pt}]  (10,270) -- (490,270) ;
\draw [color={rgb, 255:red, 208; green, 2; blue, 27 }  ,draw opacity=1 ][line width=1.5]    (110,120) -- (90,120) ;
\draw [color={rgb, 255:red, 208; green, 2; blue, 27 }  ,draw opacity=1 ][line width=1.5]    (110,180) -- (90,180) ;
\draw [color={rgb, 255:red, 208; green, 2; blue, 27 }  ,draw opacity=1 ][line width=1.5]    (110,60) -- (90,60) ;
\draw [color={rgb, 255:red, 208; green, 2; blue, 27 }  ,draw opacity=1 ][line width=1.5]    (90,60) -- (90,240) ;
\draw [color={rgb, 255:red, 208; green, 2; blue, 27 }  ,draw opacity=1 ][line width=1.5]    (90,240) -- (210,240) ;
\draw [color={rgb, 255:red, 208; green, 2; blue, 27 }  ,draw opacity=1 ]   (248.5,280) -- (248.5,260)(251.5,280) -- (251.5,260) ;
\draw [color={rgb, 255:red, 208; green, 2; blue, 27 }  ,draw opacity=1 ]   (348.5,280) -- (348.5,241.5) -- (290,241.5)(351.5,280) -- (351.5,238.5) -- (290,238.5) ;
\draw [color={rgb, 255:red, 74; green, 144; blue, 226 }  ,draw opacity=1 ]   (250,160) -- (250,140) ;
\draw [color={rgb, 255:red, 139; green, 87; blue, 42 }  ,draw opacity=1 ]   (310,180) -- (290,180) ;
\draw [color={rgb, 255:red, 139; green, 87; blue, 42 }  ,draw opacity=1 ]   (450,220) -- (350,220) ;
\draw [color={rgb, 255:red, 139; green, 87; blue, 42 }  ,draw opacity=1 ]   (450,280) -- (450,220) ;
\draw [color={rgb, 255:red, 74; green, 144; blue, 226 }  ,draw opacity=1 ]   (250,100) -- (250,80) ;
\draw [color={rgb, 255:red, 74; green, 144; blue, 226 }  ,draw opacity=1 ]   (350,110) -- (350,80) ;
\draw [color={rgb, 255:red, 74; green, 144; blue, 226 }  ,draw opacity=1 ]   (290,110) -- (350,110) ;
\draw [color={rgb, 255:red, 74; green, 144; blue, 226 }  ,draw opacity=1 ]   (200,130) -- (210,130) ;
\draw [color={rgb, 255:red, 74; green, 144; blue, 226 }  ,draw opacity=1 ]   (200,130) -- (200,230) ;
\draw [color={rgb, 255:red, 74; green, 144; blue, 226 }  ,draw opacity=1 ]   (200,230) -- (210,230) ;
\draw   (210,161) -- (290,161) -- (290,201) -- (210,201) -- cycle ;

\draw [color={rgb, 255:red, 139; green, 87; blue, 42 }  ,draw opacity=1 ]   (130,289.5) -- (110,289.5) ;
\draw [color={rgb, 255:red, 208; green, 2; blue, 27 }  ,draw opacity=1 ][line width=1.5]    (110,329.5) -- (130,329.5) ;
\draw [color={rgb, 255:red, 74; green, 144; blue, 226 }  ,draw opacity=1 ]   (110,309.5) -- (130,309.5) ;
\draw  [color={rgb, 255:red, 155; green, 155; blue, 155 }  ,draw opacity=1 ] (100,279.5) -- (190,279.5) -- (190,339.5) -- (100,339.5) -- cycle ;

\draw   (310,160) -- (390,160) -- (390,200) -- (310,200) -- cycle ;

\draw [color={rgb, 255:red, 139; green, 87; blue, 42 }  ,draw opacity=1 ]   (350,200) -- (350,220) ;
\draw [color={rgb, 255:red, 139; green, 87; blue, 42 }  ,draw opacity=1 ]   (390,180) -- (410,180) ;
\draw   (410,101) -- (490,101) -- (490,141) -- (410,141) -- cycle ;

\draw [color={rgb, 255:red, 74; green, 144; blue, 226 }  ,draw opacity=1 ]   (290,130) -- (410,130) ;
\draw [color={rgb, 255:red, 208; green, 2; blue, 27 }  ,draw opacity=1 ][line width=1.5]    (210,180) -- (190,180) ;
\draw [color={rgb, 255:red, 208; green, 2; blue, 27 }  ,draw opacity=1 ][line width=1.5]    (210,120) -- (190,120) ;

\draw (250,120) node   [align=left] {MCU};
\draw (250,239) node   [align=left] {\begin{minipage}[lt]{42.42pt}\setlength\topsep{0pt}
\begin{center}
4x Motor\\Driver
\end{center}

\end{minipage}};
\draw (450,300) node   [align=left] {\begin{minipage}[lt]{44.14pt}\setlength\topsep{0pt}
\begin{center}
Load cell
\end{center}

\end{minipage}};
\draw (450,181) node   [align=left] {\begin{minipage}[lt]{43.55pt}\setlength\topsep{0pt}
\begin{center}
Pressure\\Gauge
\end{center}

\end{minipage}};
\draw (250,301) node   [align=left] {\begin{minipage}[lt]{34.48pt}\setlength\topsep{0pt}
\begin{center}
Pumps\\P1, P2
\end{center}

\end{minipage}};
\draw (350,301) node   [align=left] {\begin{minipage}[lt]{33.34pt}\setlength\topsep{0pt}
\begin{center}
Valves\\V1, V2
\end{center}

\end{minipage}};
\draw (250,60) node   [align=left] {\begin{minipage}[lt]{23.69pt}\setlength\topsep{0pt}
\begin{center}
USB
\end{center}

\end{minipage}};
\draw (350,60) node   [align=left] {\begin{minipage}[lt]{30.31pt}\setlength\topsep{0pt}
\begin{center}
UART
\end{center}

\end{minipage}};
\draw (150,179) node   [align=left] {\begin{minipage}[lt]{35.02pt}\setlength\topsep{0pt}
\begin{center}
DC/DC\\5V
\end{center}

\end{minipage}};
\draw (150,60) node   [align=left] {\begin{minipage}[lt]{48.81pt}\setlength\topsep{0pt}
\begin{center}
BATTERY
\end{center}

\end{minipage}};
\draw (150,121) node   [align=left] {\begin{minipage}[lt]{35.02pt}\setlength\topsep{0pt}
\begin{center}
DC/DC\\3V3
\end{center}

\end{minipage}};
\draw (11,52) node [anchor=north west][inner sep=0.75pt]   [align=left] {Interfaces};
\draw (11,291) node [anchor=north west][inner sep=0.75pt]   [align=left] {Peripherals};
\draw (11,142) node [anchor=north west][inner sep=0.75pt]   [align=left] {On-board};
\draw (250,181) node   [align=left] {\begin{minipage}[lt]{33.83pt}\setlength\topsep{0pt}
\begin{center}
{\small ADC}\\{\small OpAmp}
\end{center}

\end{minipage}};
\draw (350,180) node   [align=left] {\begin{minipage}[lt]{34.5pt}\setlength\topsep{0pt}
\begin{center}
Analog\\Mux
\end{center}

\end{minipage}};
\draw (132,289.5) node [anchor=west] [inner sep=0.75pt]   [align=left] {Analog};
\draw (132,329.5) node [anchor=west] [inner sep=0.75pt]   [align=left] {Power};
\draw (132,309.5) node [anchor=west] [inner sep=0.75pt]   [align=left] {Digital};
\draw (450,121) node   [align=left] {\begin{minipage}[lt]{37.32pt}\setlength\topsep{0pt}
\begin{center}
Tactile\\Buttons
\end{center}

\end{minipage}};

\end{tikzpicture}%
    }
    \caption{Sensors and control topology. The system features a quad-channel current-chopping motor driver for the pumps P1, P2, and valves V1, V2. The sensors (pressure gauge and load cell) are interfaced through a multiplexer into a differential operational amplifier + ADC. Communication with the main microcontroller is assured via USB or UART. The power rails are generated from the battery using cascading linear voltage regulators.}
    \label{fig:sensors-and-controller}
\end{figure}

The system depicted in the block diagram in \cref{fig:sensors-and-controller} is implemented on a single, completely custom $\SI{47}{\milli\meter}\times\SI{47}{\milli\meter}$ controller board which is shown in \cref{fig:gripper-asm}. 
It is designed to work and integrate easily with common UAV hardware. 
As such, it can be powered directly from the main power bus of the drone.
Furthermore, it features USB and UART serial ports for communication with an autopilot or an off-board computer.

At the heart of the controller is an ultra-low power STM32L1 microcontroller that does the logic processing, collection/processing of the sensor data, the communication with the off-board peripherals, and the control of the quad-channel motor driver that powers the pneumatic hardware.

Due to the low power requirements of the controller (less than \SI{50}{\milli\ampere} at \SI{12}{\volt}) we have favored linear DC/DC regulators over switching converters for the $\SI{5}{\volt}$ and $\SI{3.3}{\volt}$ rails as the latter greatly increase the design complexity and cost. 
The output stage (valves and pumps) is directly powered by the main power bus. 
Current chopping motor drivers ensure that each actuator operates at its nominal operating point regardless of the bus voltage.

The load cell and the onboard air pressure sensor provide the required data for the system to monitor itself and to work autonomously.
The processed sensor readings are exposed via serial to enable more advanced applications.
Such applications include activation force tracking, the possibility of feeding back the weight of the grasped payload to the controller as a known disturbance, and the detection of a successful or unsuccessful grasp after takeoff based on the load cell readings.
We refer to the measured force $F_m$ in gram-force '\SI{}{\gramforce}' and the measured pressure as $P$ in '\SI{}{\kilo\pascal}'.
This enables applications such as controlling the activation force and also empowers the internal logic to control the pressure inside the membrane and to prevent conditions such as membrane rupture due to over-pressure and to assure a consistent air pressure while approaching the payload.

We define two pressure thresholds, namely $P_{min}=\SI{-21}{\kilo\pascal}$, the lower trigger point, and $P_{max}=\SI{0.5}{\kilo\pascal}$ the upper trigger point.
Those trigger points are used to switch reliably between the 'closed' and 'opened' states of the gripper.
In particular, $P \geq P_{max}$ signals that the membrane is full and any additional air would stretch the membrane (consequently increasing the internal pressure). 
$P \leq P_{min}$ signals that a vacuum is established and thus the gripper is considered 'closed'.

The firmware on the MCU is making use of \textit{FreeRTOS}, running two tasks using preemptive multitasking as shown in \cref{alg:tasks}. 
Task 1 handles sensors and actuation, and task 2 handles serial communication.
Inter-task communication takes place over thread-safe FIFO queues. 
For the underlying state machine (automaton), we direct the reader to the \hyperref[sec:appendix]{Appendix}.

\begin{algorithm}[tb]
    \caption{FreeRTOS, sensor acquisition and actuation}
    \label{alg:tasks}
    \begin{algorithmic}
        \Procedure{task 1: sensors and actuation}{}
            \State let $k_{gr}$ be the current gripper state
            \State let $S(k_{gr})$ be the automaton in \cref{fig:automaton}
            \State let $f_1$, $f_2$ be lowpass FIR filters
            \Loop
            \State collect push button states $\mathbf{u_{bt}}$
            \State fetch raw sensor data $P^*$, $F_m^*$
            \State process sensor data $P \gets f_1(P^*)$, $F_m \gets f_2(F_m^*)$
            \State create state vector $\mathbf{q} \gets (t, k_{gr}, P, F_m)$
            \State process automaton $\mathbf{u_a} \gets S(\mathbf{q}, \mathbf{u})$)
            \State apply $\mathbf{u_a}$ to actuators 
            \EndLoop
        \EndProcedure
    \end{algorithmic}
    \begin{algorithmic}
        \Procedure{task 2: communication}{}
            \Loop
            \State outbound communication, send $\mathbf{q}$
            \State inbound communication, receive $\mathbf{u_{usr}}$
            \State create command vector $\mathbf{u} \gets (\mathbf{u_{usr}}, \mathbf{u_{bt}})$
            \EndLoop
        \EndProcedure
    \end{algorithmic}
\end{algorithm}

\subsection{Grasping Procedure}
\label{sec:grasping-procedure}

\begin{figure}
    \centering
    \resizebox{0.7\linewidth}{!}{
    \input{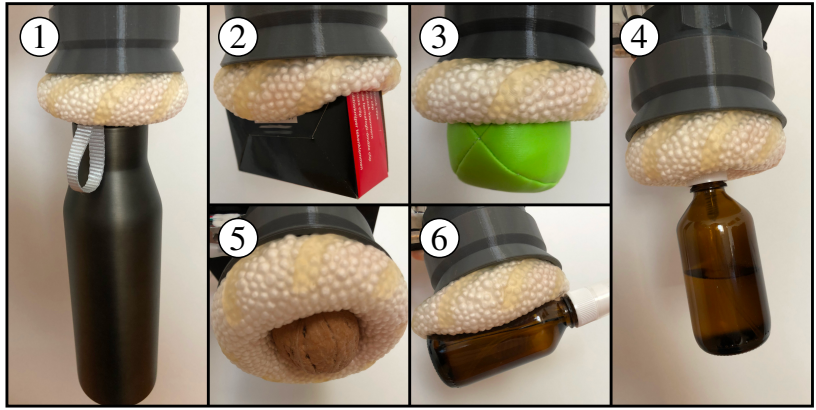}
    }
    \caption{TRIGGER grasping a variety of objects, showing its universal grasping ability. \MARKERCIRCLE{1} Empty aluminium bottle, \MARKERCIRCLE{2} cardboard box, \MARKERCIRCLE{3} semi-soft ball, \MARKERCIRCLE{4} filled glass flask hold from the small top cover, \MARKERCIRCLE{5} walnut, \MARKERCIRCLE{6} filled glass flask held from the smooth, slippery side.}
    \label{fig:grasped-objects} 
\end{figure}

\begin{figure}
    \centering
    \resizebox{0.6\linewidth}{!}{%
    \includegraphics[]{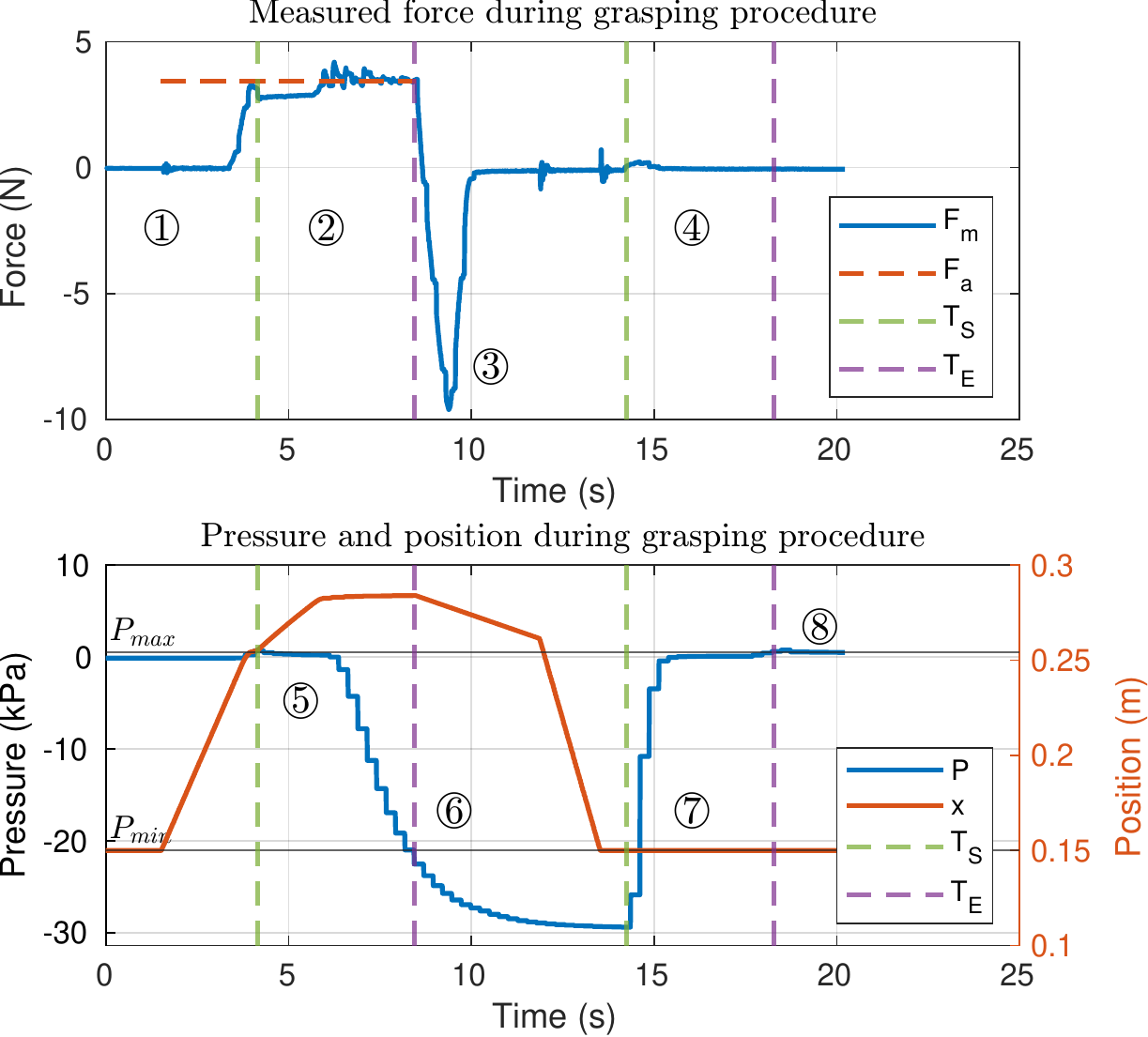}%
    }
    \caption{Grasping procedure. \MARKERCIRCLE{1} Approach, \MARKERCIRCLE{2} evacuation phase where the air is pumped out of the membrane, \MARKERCIRCLE{3} retraction phase where the peek marks the maximum holding force, \MARKERCIRCLE{4} reset of the gripper by pumping air into the membrane, \MARKERCIRCLE{5} pressure starts dropping once the volume reaches the minimum, \MARKERCIRCLE{6} end of closing procedure triggered by $P \leq P_{min}$, \MARKERCIRCLE{7} pressure rises as air is pumped in, \MARKERCIRCLE{8} end of opening procedure triggered by $P \geq P_{max}$. $T_S$ and $T_E$ mark the start and the end of the transition between open and close and vice-versa.}
    \label{fig:procedure} 
\end{figure}

Although usage of our UG by hand is straightforward and allows grasping of a variety of shapes and materials (see \cref{fig:grasped-objects}), a defined grasping procedure is required for our aerial platform such that successful grasps can be achieved without relying on human intuition.
Typically this procedure consists of four main steps (\cref{fig:procedure}):
\begin{enumerate}
    \item The grasp starts by pushing the fluidized gripper against the payload. 
    Doing so elastically deforms the membrane and the filler material flows freely, distributing itself around the payload. 
    At this point, valves V1 and V2 are still closed such that the free volume remains unchanged. 
    The evacuation phase is then triggered once the measured force reaches the desired activation force, i.e., $F_{m} \geq F_a$. 
    \item Evacuating the air out of the membrane takes a couple of seconds (governed by the flow rate of the pumps). 
    During that period, the membrane shrinks, and the contact force drops in response to that unless the gripper is further moved toward the payload. 
    In the context of low activation forces, it is essential to keep good contact with the payload. 
    Failure to do so will lead to a poor or unsuccessful grasp as the filler hardens without properly surrounding the payload. 
    We thus track the nominal activation force during this interval. 
    Other publications in this field usually avoid this step by pushing the gripper with a very high force into the payload, e.g., with \SI{17}{\newton} as seen in \cite{GomezPaccapelo2021}, which is, however, not possible with most small to medium aerial systems.
    \item Once the membrane's internal pressure satisfies $P \leq P_{min}$ (vacuum), the grasping procedure is considered completed. The gripper is then retracted from the payload (here, at a constant velocity). 
    Since the payload is fixated on the support and cannot be lifted, a negative force is measured corresponding to the holding force $F_h$. 
    In practice, with the gripper mounted on a UAV, instead of the holding force, the actual weight of the lifted payload would be measured, which could be fed back into the autopilot.
    \item Releasing the payload (i.e., opening resp. resetting the gripper) is achieved by pumping air into the membrane until $P \geq P_{max}$ is reached, then closing valves V1 and V2. 
    The gripper is now ready to grasp the next object.
\end{enumerate}

The activation force $F_{a}$ is an essential quantity for a successful grasping operation.
An insufficient activation force causes the grasping operation to fail.
On the other hand, choosing $F_{a}$ too high may destabilize the aerial platform and also cause damage to the force sensor and membrane. 
Therefore, the optimal activation force has to be chosen sufficiently high not to sacrifice performance but also as low as possible to minimize the impact on the aerial platform.

\section{Experiments}
\label{sec:experiments}
In the following section, we introduce our experimental setup as well as the experiments to, \textit{first}, determine the optimal minimal activation force for the grasping procedure, and \textit{second}, to access the influence of the silicone additive \textit{deadener} on the grasping performance.

We also use this experimental setup to collect data regarding the stiffness of the UG in various states with the goal to create a simple simulation model resp. contact model for common robotics simulators (\cref{sec:digital-twin}). 

\subsection{Experimental Setup}

\begin{figure}
    \centering
    \resizebox{0.5\linewidth}{!}{%
    \input{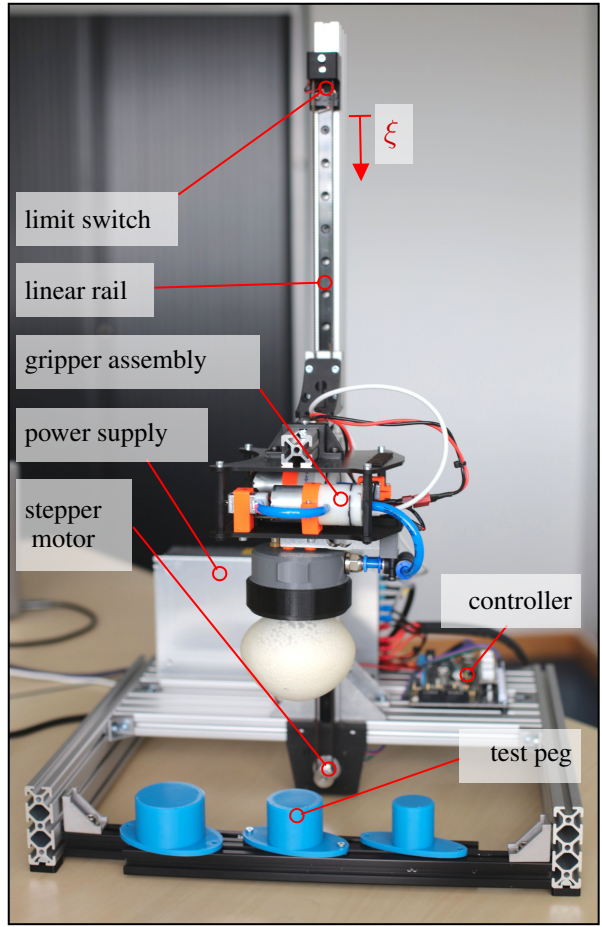}%
    }
    \caption{Experimental jig. The UG is attached to a belt-driven linear rail, moving the gripper assembly into contact with the test peg (blue). The test peg is a simple cylindrical object with no features allowing geometric interlocking.}
    \label{fig:jig} 
\end{figure}

Our experimental jig used for benchmarking is featured in \cref{fig:jig}. 
It consists of a \SI{12}{\volt} supply and a single belt-driven linear rail (capable of fast movements) moving the entire gripper assembly fixated to the horizontal beam. The stepper motor is powered and controlled by an off-the-shelf \textit{3D printer board}. 
The jig does not feature any sensors since they are already integrated into the gripper itself.

Velocity commands $u_{feed}$ as well as position commands (G-codes) are accepted by our custom firmware on the jig's controller. 
The current position and velocity are sent back to the host computer for the purpose of logging.
Likewise, the gripper's controller sends back its current state and sensor data (force, pressure, and input voltage) while also accepting state transition commands (open/close).

Cylindrical test pegs (blue) are representing our payloads. 
They are fixated at the bottom of the $\xi$-axis, such that they cannot be lifted.
The center of the pegs is aligned with the center of the membrane.
They have no features that would allow for geometric interlocking. 
As such, the results concerning $F_h$ can be seen as a worst-case scenario.

All of our experiments require tracking of a nominal activation force. As such, given a nominal force $F_d$ and the measured force $F_{m}$, we want to drive the error $e_f$ close to zero:
\begin{align}
    e_{f} = F_{d} - F_{m}
\end{align}
The input $u_{feed}$ is the commanded linear velocity of the sled. The system itself acts as an integrator since $F_{m}$ is a function of the position, and as such $e_f$ is guaranteed to be driven close to zero by the simple proportional control law
\begin{align}
    u_{feed} = K_p e_f,
\end{align}
where $K_p$ is a positive constant.

\subsection{Minimum Activation Force}
The activation force is a crucial factor for a successful grasp. 
However, in the context of aerial manipulation, where the base of the gripper is floating, this becomes an even more important factor as any external forces have the potential to cause stability issues. 
We would like to point out the following key aspects:
\begin{itemize}
    \item An aerial platform is severely limited in the amount of force it can apply to its environment before hitting its stability margins.
    \item Under some circumstances, e.g., when the target is poorly supported, it is impossible to apply a substantial activation force.
    \item The need for a (large) activation force can thus be seen as a net disadvantage of these types of grippers. Reducing it is therefore considered beneficial.
\end{itemize}

Recent research reported that there is a monotonic relation between the activation force and the resulting holding force \cite{GomezPaccapelo2021}. 
An older study found that after reaching a certain threshold, the holding force stays constant \cite{Kapadia2012}.
Our results confirm the findings of both studies and indicate that this threshold depends on the fill ratio of the membrane.

In this study, we will focus on small activation forces $F_a < \SI{650}{\gramforce}$ as they are the most useful for aerial grasping. 
We thus conduct six test series with nominal activation forces ranging from \SI{150}{\gramforce} to \SI{650}{\gramforce} for fill ratios of $66\%$ and $90\%$.
Each test series follows the grasping procedure described in \cref{sec:grasping-procedure} and is repeated eight times using the \diameter\SI{40}{\milli\meter} peg. 
For this and all subsequent experiments, we have chosen $K_p=6$ as a compromise to achieve reasonable force tracking while the membrane is still soft and to reduce overshoots when the membrane hardens at the end of the evacuation phase.

\begin{figure}
    \centering
    \includegraphics[width=0.6\linewidth]{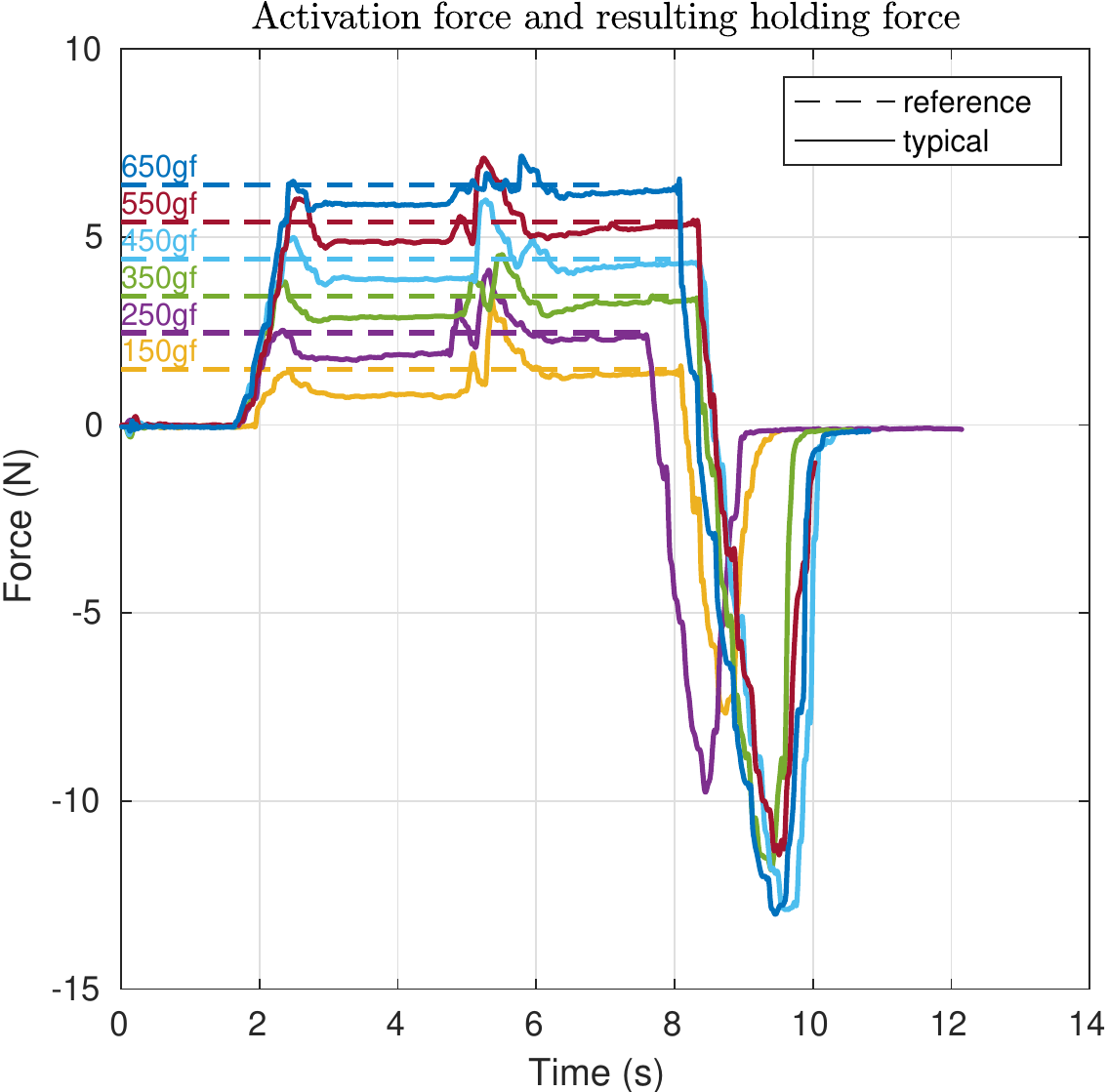}
    \caption{Activation and holding force for various nominal activation forces from \SI{150}{\gramforce} to  \SI{650}{\gramforce}. The nominal activation force is tracked during the hole evacuation phase (\SI{2}{\second} to \SI{8}{\second}). The maximum holding force is then measured during the retraction phase (\SI{8}{\second} to \SI{10}{\second}), where it shows up as the peak negative force.}
    \label{fig:activation-force-holding-force-time} 
\end{figure}

\begin{figure}
    \centering
    \includegraphics[width=0.6\linewidth]{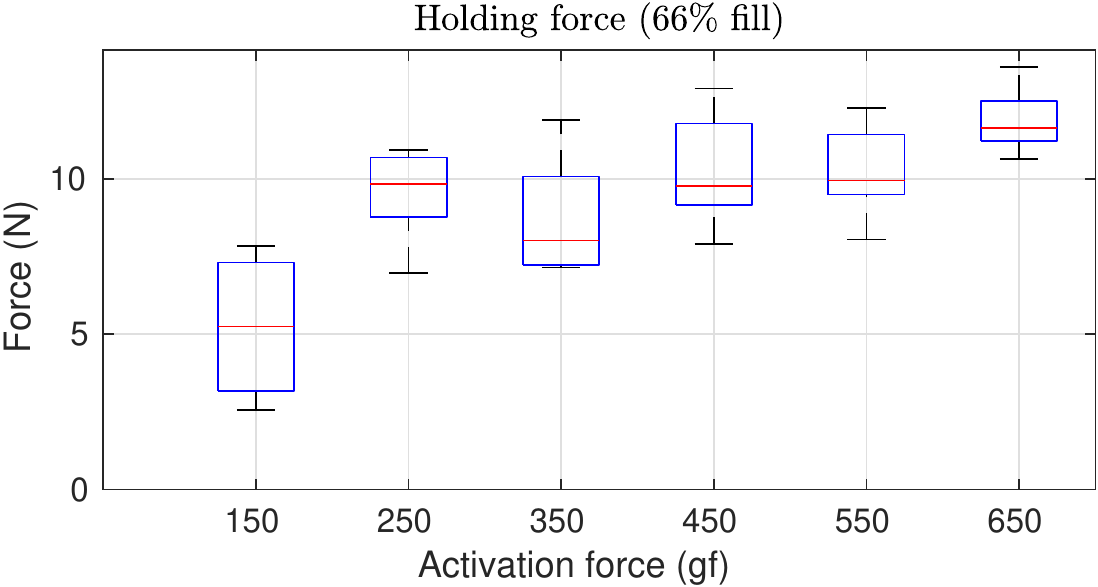}
    \caption{Holding force in relation to the nominal activation force, with a fill ratio of \SI{66}{\percent}. The holding force stays constant after reaching the threshold of \SI{250}{\gramforce}.}
    \label{fig:hold-force-holding-force-boxplot} 
\end{figure}

\begin{figure}
    \centering
    \includegraphics[width=0.6\linewidth]{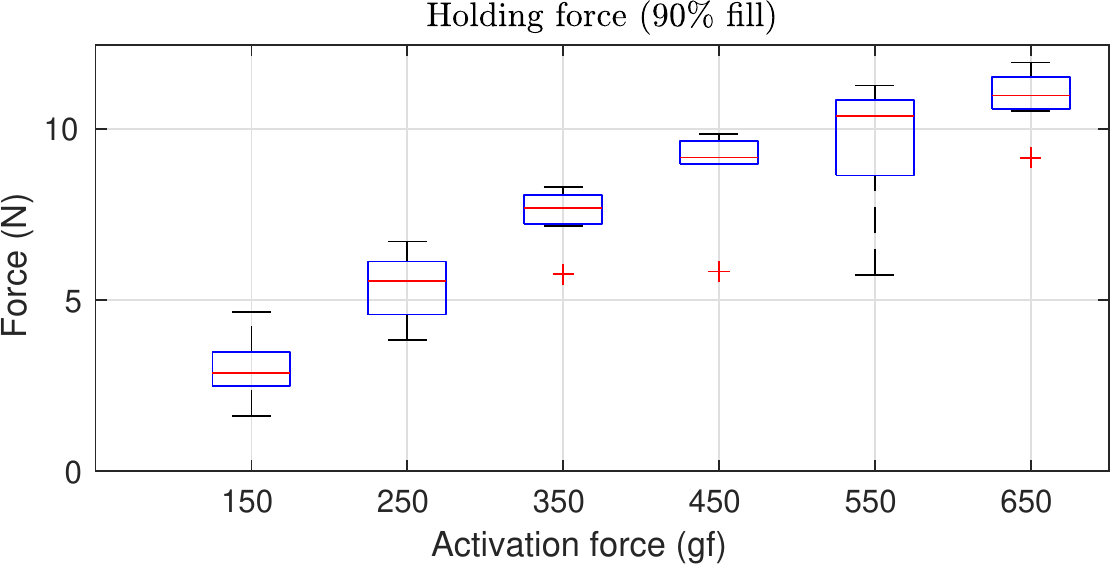}
    \caption{Holding force in relation to the nominal activation force, with a higher fill ratio of \SI{90}{\percent}. The holding force increases monotonically with the activation force.}
    \label{fig:hold-force-holding-force-boxplot-high-infill} 
\end{figure}

The results of this study are shown in \cref{fig:activation-force-holding-force-time,fig:hold-force-holding-force-boxplot-high-infill}. 
The data shows that TRIGGER can reach a maximum holding force $F_{h,max}$ of about \SI{10}{\newton} for the test peg with $D=\diameter\SI{40}{\milli\meter}$ and without geometric interlocking. 
The evacuation period typically takes $T_{SE}=T_E-T_S=\SI{4.3}{\second}$ (\cref{fig:procedure}).
Furthermore, there is a clear relationship between the fill ratio and the minimal activation forces required to reach the maximal holding force, as illustrated in  \cref{fig:hold-force-holding-force-boxplot,fig:hold-force-holding-force-boxplot-high-infill}.
Lower fill ratios generally reduce the required minimum activation force.
The gripper with the membrane having the lower fill ratio of $66\%$ reaches $F_{h,max}$ with an activation force of only \SI{250}{\gramforce} (see \cref{fig:hold-force-holding-force-boxplot}).
Increasing the activation force beyond that threshold does not significantly increase the resulting holding force.
In case of the higher infill ratio of $90\%$, \SI{650}{\gramforce} are required to get the same holding force (see \cref{fig:hold-force-holding-force-boxplot-high-infill}).
Higher fill ratios naturally come with a smaller free volume, i.e., the volume in which the grains can move freely.
In turn, the mobility of filler particles is impaired, which then requires a higher effort to redistribute the filler within the membrane during contact.
This manifests in a monotonic relationship between the activation force and the holding force. 
Therefore the minimum activation force increases with the fill ratio.

We thus conclude that an activation force $F_a \geq \SI{250}{\gramforce}$ is adequate for successful grasping without compromising the holding capability of the gripper. 
We consider this low enough to work on a wide range of UAVs without significantly impacting the stability of the aerial system. 
Furthermore, we conclude that lower fill ratios (in the $60\%$ range) are preferable since they require lower activation forces.

\subsection{Deadener (Additive)}
\textit{Deadener} (also called \textit{slacker}) is an additive that is added to the silicone during the mixing process. 
It alters the physical properties of the cured silicone by increasing its softness and stickiness. 

Herein, we measured the effect of adding \SI{0}{\percent} to \SI{15}{\percent} of deadener (by weight) to the mixture.
At around \SI{15}{\percent} deadener, the membrane reached a consistency similar to chewing gum.
Further increasing the percentage was thus deemed impractical.

\begin{figure}
    \centering
    \resizebox{0.6\linewidth}{!}{
    \includegraphics[]{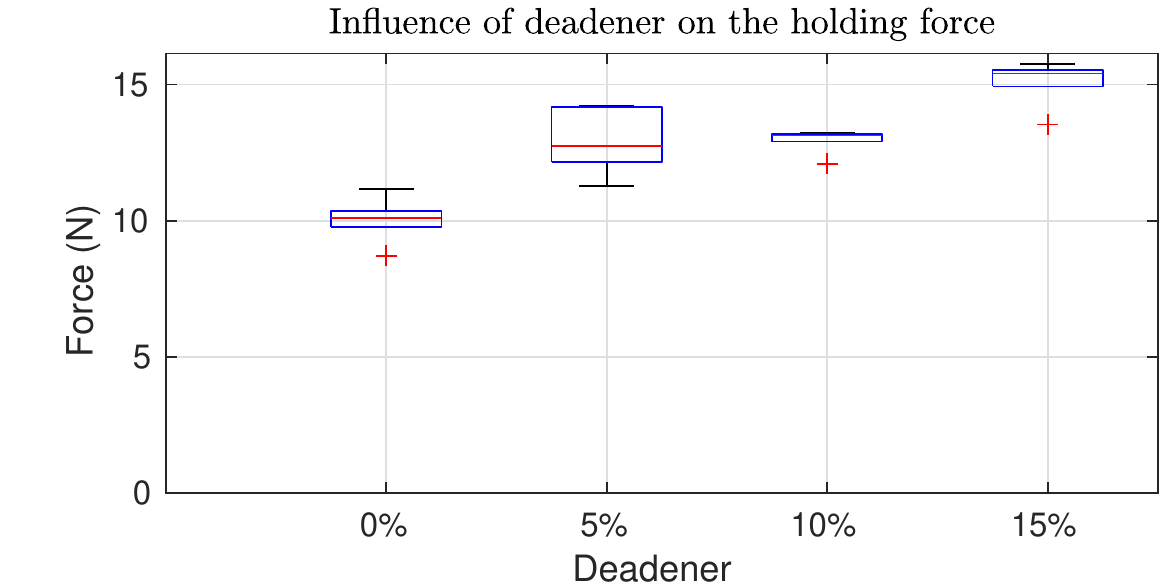}
    }
    \caption{Holding force for various percentages of deadener added to the silicone mixture. Increasing the amount of deadener turns the membrane softer and stickier, increasing the holding force.}
    \label{fig:deadener} 
\end{figure}

For the test series, we created four membranes with \SI{0}{\percent}, \SI{5}{\percent}, \SI{10}{\percent} and \SI{15}{\percent} deadener. 
We repeated our holding force test 6 times for each of the membranes, with an activation force of \SI{350}{\gramforce}, on a \diameter\SI{40}{\milli\meter} test peg.
The results are shown in \cref{fig:deadener}.
Starting with no deadener, we reach the expected holding force of around \SI{10.1}{\newton}. 
A median of \SI{12.8}{\newton} and \SI{13.1}{\newton} was measured for \SI{5}{\percent} and \SI{10}{\percent} deadener, respectively. 
Further increasing it to \SI{15}{\percent}, resulting in a median holding force of \SI{15.4}{\newton}, a significant increase of \SI{52}{\percent} compared to no deadener.

The increase in performance can be explained as follows. 
On one hand, the additive turns the membrane slightly tacky, thus increasing the friction coefficient between the membrane and the test peg. 
On the other hand, due to the increased softness, the membrane's ability to conform to the test peg's shape is improved, resulting in a larger contact surface.
Both of them cause the contribution of the friction $F_R$ to increase, which results in a greater $F_h$.
It should be noted that the stickiness is a temporary effect and dwindles over time as the silicone ages and dust and dirt accumulate on the membrane's surface.
For a more lasting effect, a clean environment is thus required. 
Alternatively, a periodic replacement/renewal of the membrane module (which is fully supported by our design, see \cref{fig:gripper-asm}) should be considered.

\section{Modeling}
\label{sec:digital-twin}
In the context of aerial manipulation, the UG exhibits challenging dynamic behavior as it transitions from a soft state to a jammed (almost rigid) state.
This section proposes a homologous model of TRIGGER based on observations and measurements obtained from our experiments. 

Our model will faithfully represent the following main aspects of the UG:
\begin{enumerate}
    \item the normalized free volume $\beta$ represented as a first-order system, which models the transition between the membrane's open/closed states.
    \item the membrane shrinkage $x_a \left(\beta\right)$ caused by the changing free air volume.
    \item the contact force contribution of the air-filled membrane represented by the compression spring $k_{air}$ as a function of the payload diameter and the normalized free volume $\beta$.
    \item the contact force contribution due to the lumped elasticity represented by the compression spring $k_{lmp}$, which takes into account the complete assembly with the gripper being jammed.
\end{enumerate}
We assume negligible damping, constant volumetric airflow and $k_{air} \ll k_{lmp}$ and that the membrane does not touch the ground during the grasping phase (satisfied for any reasonably sized payload).

\begin{figure}
    \centering
    \resizebox{0.7\linewidth}{!}{%
    {\input{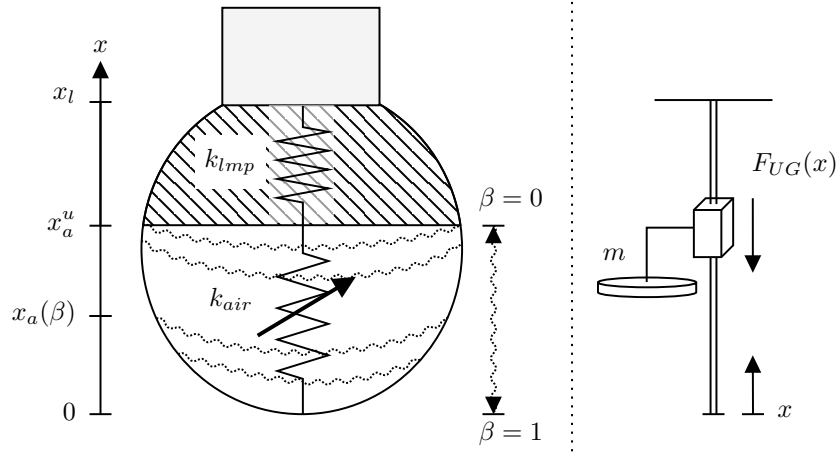}}%
    }
    \caption{The UG membrane is separated into two components: the air-filled elastic membrane and the rest of the system. Both components are modeled as compression springs $k_{air}$ and $k_{lmp}$ respectively (left). The equivalent mechanical system consists of a disk (body) with a mass $m$ attached to a prismatic joint with finite travel subjected to the joint force $F_{UG}$ (right).}
    \label{fig:contact-model} 
\end{figure}

We propose the contact model shown in \cref{fig:contact-model}, consisting of the two non-linear compression springs $k_{lmp}$ and $k_{air}$, where the latter is of variable stiffness, i.e., dependant on the size of the targeted object. 
The spring $k_{lmp}$ represents the lumped stiffness of the jammed filler material and the structural parts of the assembly. 
The spring $k_{air}$ represents the compression of the air-filled membrane during contact. 
Its stiffness is tied to many parameters, such as the effective contact area, the non-linear elastic behavior of the membrane, internal pressure, filled volume, and other factors of which most cannot be measured.
Its dynamic behavior is, to some extent, akin to an air spring, e.g., \cite{Zhu2017}. 
However, such a precise model is very hard to identify and has no practical benefits in this context.

\begin{figure}
    \centering
    \resizebox{0.6\linewidth}{!}{%
    \input{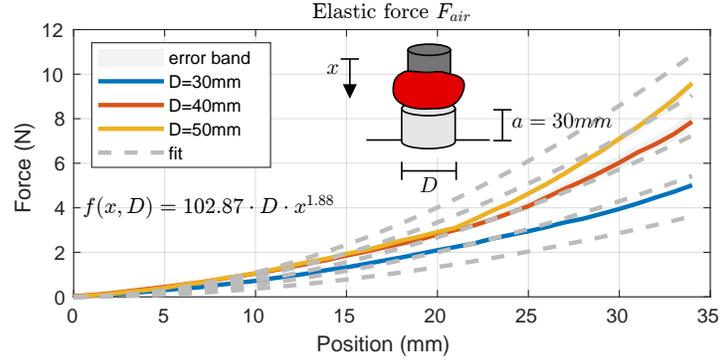}%
    }
    \caption{Elastic force $F_{air}$ of the inflated membrane during contact with different sized test pegs from \SI{30}{\milli\meter} to \SI{50}{\milli\meter}. Larger effective areas generate a higher elastic force.}
    \label{fig:air-stiffness} 
\end{figure}

\begin{figure}
    \centering
    \resizebox{0.6\linewidth}{!}{\input{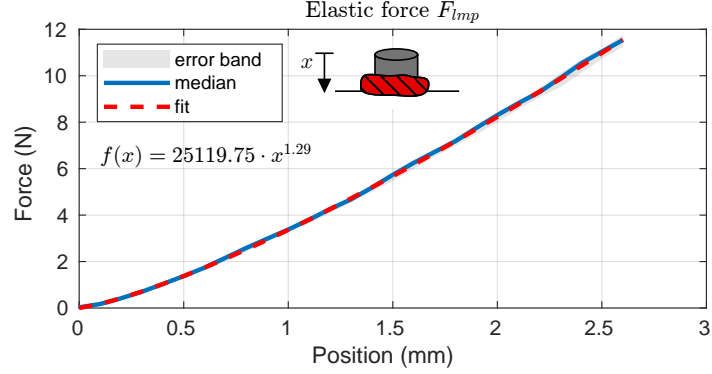}}
    \caption{Lumped elastic force $F_{lmp}$ of the jammed membrane-gripper system during contact with the test peg. The diameter of the test peg does not affect the results since the shape of the membrane does not change.}
    \label{fig:lumped-stiffness} 
\end{figure}

We employed non-linear regression analysis to identify the relation between the depth of entrance $x$, the payload/peg diameter $D \in \left(0,60\right]$ (in \SI{}{\milli\meter}) and the resulting elastic force $F_{air}$ as follows (\cref{fig:air-stiffness})
\begin{align}
\begin{split}
    F_{air}\left(x, D\right) &= a_1 D x^{a_2} \\
                             &= 102.87 \cdot D x^{1.88}.
\end{split}
\end{align}
Similarly, we identified the lumped elastic force of the system to be (\cref{fig:lumped-stiffness})
\begin{align}
\begin{split}
    F_{lmp}\left(x\right) &= a_3 x^{a_4} \\
                          &= 25119.75 \cdot x^{1.29}.
\end{split}
\end{align}
The combined elastic force $F_{ug}$ is then obtained by
\begin{align}
    F_{ug}\left(x, D\right) = F_{air}\left(H\left(x\right), D\right) + F_{lmp}\left(H\left(x - x_a\right)\right),
    \label{eq:contact-force}
\end{align}
where
\begin{align}
    H(x)=
    \begin{cases}
        0 & \text{if x < 0,} \\
        x & \text{if x > 0,}
    \end{cases}
\end{align}
accounts for the fact that the springs act in compression only.

Another key aspect represented by the model is the shrinkage of the membrane during the evacuation phase (see \cref{fig:procedure,fig:contact-model}), where the air is pumped out of the membrane, which consequently shrinks in the process. 
Herein, it is assumed that this shrinkage follows an exponential law and we designate this internal state by the letter $\beta \in [0,1]$, where $\beta=1$ indicates that the membrane is completely filled with air (without stretching it) and $\beta=0$ means that all the air is evacuated (the filler is jammed resp. in the process of getting jammed).
The transition phase is modeled as a first-order system defined as
\begin{align}
    \beta\left(s\right) = \frac{R\left(s\right)}{1+s T},
    \label{eq:ug-system-ratio}
\end{align}
where $T=\SI{63}{\percent} \cdot T_{SE} = \SI{2.8}{\second}$ is the time constant and $R\left(s\right)$ being a unit step (either 0 or 1, depending on the desired state transition).
Based on the value of $\beta$, we define three discrete gripper states such that
\begin{align}
    k_{gr}(\beta) =
    \begin{cases}
        \text{closed} & \text{if $\beta \leq \SI{1}{\percent}$} \\
        \text{opened} & \text{if $\beta \geq \SI{99}{\percent}$} \\
        \text{in transition} & \text{otherwise} \\
    \end{cases}.
\end{align}

\begin{figure}
    \centering
    \resizebox{0.6\linewidth}{!}{%
    \includegraphics[]{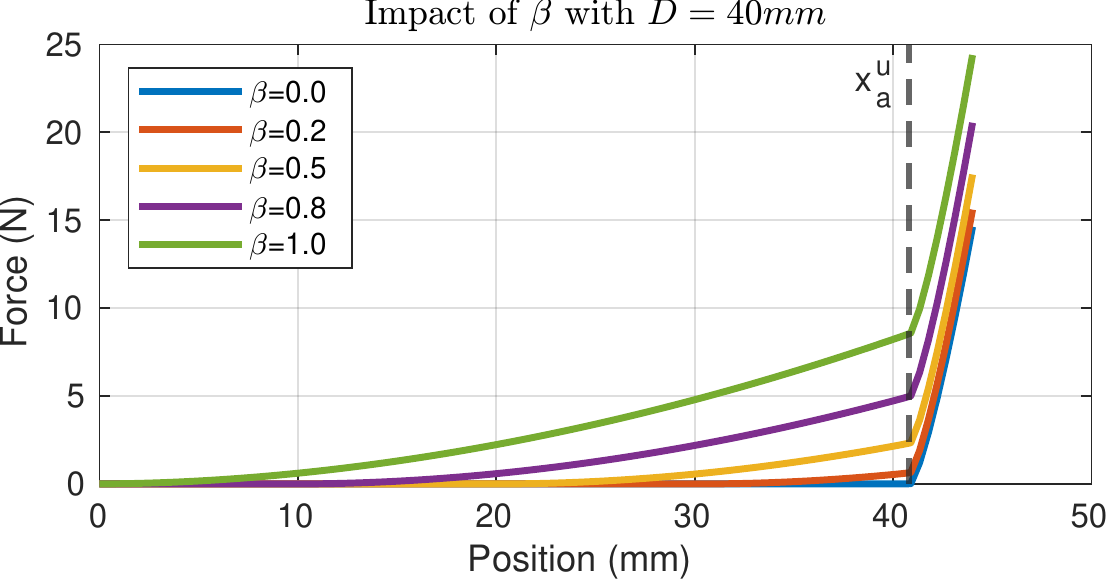}%
    }
    \caption{Contact force model for different membrane fill ratios from $\beta=0$ empty/jammed to $\beta=1$ completely filled with air. The lumped stiffness dominates starting from $x^u_{a}=\SI{40.8}{\milli\meter}$.}
    \label{fig:contact-force-digital-twin} 
\end{figure}

Next, we assume (simplify) that the free length $x^u_a$ depends on $\beta$ as described by the linear mapping
\begin{align}
    x_a \left(\beta\right) = x^u_{a} \cdot \beta,
\end{align}
where the upper bound $x^u_{a}$ is defined by the minimal volume occupied by the filler material in the jammed state. 
Herein, we identified $x^u_{a}=\SI{40.8}{\milli\meter}$ experimentally, with it being the position at which the jammed membrane first makes contact with the test peg. 
We thus conclude with the contact force model as shown in \cref{fig:contact-force-digital-twin}.

The contact model is the key component in creating a digital UG for use in a robotics simulator such as \textit{Gazebo}. 
We propose modeling the UG as a disk with a diameter of \SI{80}{\milli\meter} and a mass of $m=\SI{20}{\gram}$ (weight of filler plus membrane). That disk is attached to a prismatic joint that simulates the shrinkage of the membrane as a function of $\beta$, the intrusion of the payload into the membrane, and is subjected to a force $F_{UG}$ governed by the contact force model \eqref{eq:contact-force}.
The travel limits of the joint are defined to be in the range of $\left[0, x_l\right]$, where $x_l=\SI{60}{\milli\meter}$ is the height of the membrane (see \cref{fig:contact-model}, right).
The UG's internal state $\beta$ is governed by the system model \eqref{eq:ug-system-ratio}.

As for the grasping part, we suggest considering a grasp as successful as long as an activation force greater than \SI{250}{\gramforce} is maintained with the payload during the entire evacuation phase. 
A successful grasp should then establish a rigid connection between the gripper and the payload motivated by the jamming (hardening) of the filler material.
That connection should be removed if $F_m \geq F_h$ or as a result of the state transition $\beta > \epsilon > 0$, where $\epsilon$ is a very small value.

Another application of the contact model \eqref{eq:contact-force} could be the estimation of payload size with the help of characteristic curves of the elastic forces identified during the initial contact, as shown in \cref{fig:air-stiffness}. 
However, this requires precise measurements of the position, which may not be available in real-world conditions.

\section{Aerial Application}
\label{sec:aerial-application}
This section briefly presents a pick-and-release manipulation task with the UG attached to a UAV.
The overall goal of this experiment is to validate the fundamental concept shown in \cref{fig:concept}.

Our proof-of-concept platform based on the frame of the \textit{AscTec Firefly} features a \textit{Raspberry Pi 3} with a \textit{Navio 2} running the \textit{PX4} autopilot. 
We created a custom \textit{PX4} firmware module to communicate with the gripper and to link one of the remote control channels to trigger its opening resp, closing state transition.
Herein, the UAV is carefully manually piloted in position control mode, relying on human intuition to keep a sufficient amount of activation force.

\begin{figure*}
    \centering
    \resizebox{1.0\linewidth}{!}{%
    \input{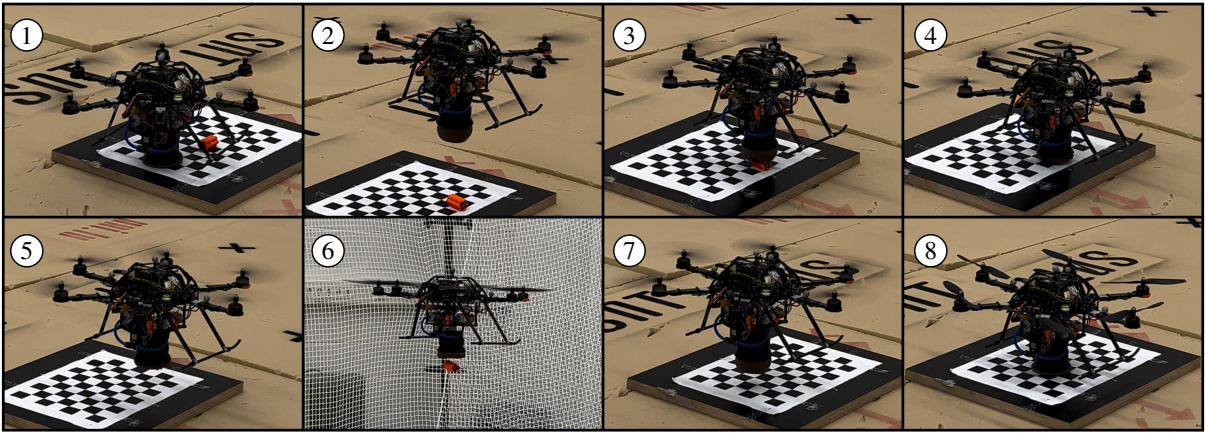}%
    }
    \caption{Aerial grasping application. \MARKERCIRCLE{1} The UAV rests on the gripper, ready for takeoff. \MARKERCIRCLE{2} The UAV approaches the payload (orange). \MARKERCIRCLE{3} The UAV gets in contact with the payload. \MARKERCIRCLE{4} The air is pumped out of the gripper and the payload is gripped. \MARKERCIRCLE{5} Takeoff with the payload. \MARKERCIRCLE{6} Air is pumped into the gripper, which releases the payload. \MARKERCIRCLE{7} The UAV approaches the landing platform and \MARKERCIRCLE{8} lands on the gripper (the landing gear does not touch the ground).}
    \label{fig:aerial-application} 
\end{figure*}

The whole experiment is pictured in \cref{fig:aerial-application} and the supplementary video is available online \footnote{Supplementary video: \href{https://youtu.be/Az5bXnZUNlY}{https://youtu.be/Az5bXnZUNlY}}.
The task of the UAV is to take off, grab the payload (orange), and drop it in the drop-off area. 
During the setup of the experiment, the UAV is manually placed on the checkerboard. 
The gripper is then closed using the push buttons on the controller board.
The membrane then forms a flat and rigid surface for the UAV to rest on.
At that stage, the UAV is ready to take off.

After successful takeoff, the membrane of the UG is fluidized (triggered by the remote) and the payload is approached. 
Ideally, the membrane would hit the payload dead center, but the UG is fairly tolerant to positional errors. 
After the activation force crosses a threshold of \SI{250}{\gramforce}, the evacuation phase is automatically triggered and the filler material hardens, creating a firm grasp on the payload. 
Now, the drone is piloted to the drop-off zone, where it releases its payload by fluidizing the gripper, triggered via the remote control.

Lastly, the UAV is piloted back to the checkerboard, where it safely lands on the UG. Notice that the landing gear, although present, never touches the ground. 
It was kept for safety reasons only.

\section{Discussion}
\label{sec:discussion}
The interest in developing soft grippers for aerial vehicles stems from the fact that by leveraging the properties of soft materials, soft grippers are a natural match for aerial grasping.
In contrast to their rigid counterparts, soft grippers are tolerant toward unknown object geometries and surfaces and do not require high positional accuracy for successful grasps.
By developing a lightweight soft jamming universal gripper attached to a UAV we further advanced the potential of soft aerial grasping. 
Compared to available soft grippers for aerial grasping, the developed system exhibits several distinguishing characteristics.

\textit{First}, the developed gripper is highly integrated and modular. 
The tight integration of the electronics, software, sensors and mechanics leads to significant weight savings and enables a well-defined grasping procedure that can be automated for use in autonomous systems.
The modularity not only helps in iterating the design rapidly, but it also addresses some concerns typically associated with UGs.
By having an explicit interface between the grasping part (membrane module) and the supporting hardware, we assure that the membrane is quick and easy (toolless) to swap in case of damage.
Other types of pneumatic grippers could also make use of this interface, e.g., suction cups.
Thanks to the specific characteristics of our UG's construction, it is particularly well suited for aerial vehicles, comparable to the typical multi-fingered soft grippers, but with some unique features (e.g., omnidirectionality, or the ability to use as landing gear).

\textit{Second}, TRIGGER is omnidirectional in contrast to other available soft aerial grasping systems (e.g., claws), which are sensitive to the angle the payload is approached.
The same applies to lateral position errors, where the UG tolerates displacements as large as \SI{60}{\percent} of its diameter.
This relaxes the requirements in terms of necessary grasping accuracy, which is especially advantageous for aerial systems that are subjected to external disturbances (e.g., wind gusts, ground effect) and sensor inaccuracies.
During contact, the UAV retains most of its degrees of freedom due to the gripper's elasticity.
As such, it can still rotate (pitch and roll) and therefore preserve hover conditions, but at the same time, the translational degrees of freedom are soft-locked by the friction between the payload and the gripper.
This would address one concern associated with soft finger grippers. 
The authors of \cite{Mishra2018} stated that during their experiment, the multicopter had to land on the ground due to ground effects and the resulting lack of precise position control.
During our aerial experiment, we did not observe such a problem as the UAV passively stayed locked in place during the grasp.

\textit{Third}, unlike soft fingers, our UG forms a rigid-like flat surface once a vacuum is established and thus enables the UAV to rest on it. 
This feature makes our manipulating UAV system exceptional as it removes the need for a dedicated landing gear that also often interferes with the attached gripper resp. the sensors required for autonomous grasping.
Moreover, using the UG as landing gear further reduces the weight of the aerial system.
The system is also able (within limits) to compensate for some terrain imperfections (e.g., slanted surfaces or small rocks), assuring optimal takeoff and land conditions.
Traditional soft finger grippers often cannot prevent the payload from moving after the grasp is established, which can create further disturbances during flight.
Contrary to our UG, which forms a system behaving much more akin to a single rigid body due to the jamming of the granular material.

\textit{Forth}, hard shocks typically associated with the impact of two bodies are problematic both from a mechanical perspective, like the risk of damage, and also from a control perspective (e.g., potential instability).
Passive mechanical compliance alleviates this problem by spreading the impact over a larger time interval.
The developed UG is completely soft during the first contact phase and is thus passively compliant and absorbs and dampens shocks.
This applies to both landing and grasping scenarios.

\textit{Fifth}, our gripper develops \SI{15}{\newton} of holding force on our test peg that do not allow for geometric interlocking and thus purely relied on friction and suction (to a much lesser extent), which is, therefore, a worst-case scenario.
As indicated in \cite{Kapadia2012}, geometric interlocking can dramatically increase the holding force.
Comparisons with other UGs are hard to make due to the lack of a standardized test procedure.
However, comparing our results with the work of \cite{Kapadia2012}, \cite{Mishra2021} and \cite{GomezPaccapelo2021}, it can be said that the measured holding force for objects without geometric interlocking is in the same neighborhood, i.e., \SI{10}{\newton}-\SI{30}{\newton}, whilst being significantly lower power (less than $\SI{10}{\watt}$ against several hundreds of watts).
Consequently, also the cycle times of our solution are longer (\SI{11}{\second} against \SI{4}{\second}) and have to be handled properly, and failure to do so will result in degraded or even unsuccessful grasps.
Therefore, in the larger context of UGs, our results indicate that high-power pumps are not strictly required.
In practice, fitting larger, heavier pumps is limited by the payload capacity of the aerial platform.

\section{Conclusion}
\label{sec:conclusion}
This work introduced TRIGGER, a novel, highly integrated, and lightweight UG with a low activation force requirement for aerial manipulation.
We experimentally validated the presented gripper and determined the optimal minimum activation force required to work reliably. We also concluded that the relation between the activation force and the resulting holding force is highly dependent on the fill ratio, i.e., lower fill ratios are preferable since they lower the required activation force.
We further investigated possible holding force improvements using a silicone additive (\textit{deadener}), which yielded a \SI{52}{\percent} higher holding force compared to the reference case without deadener.

Based on our experimental results, we developed a simulation model that faithfully represents the most relevant aspects of TRIGGER intended for numerical robotics simulators.
We also presented preliminary results showing TRIGGER attached to a hexacopter, successfully performing a pick and release task under lab conditions.

Regarding the design, we believe that significant weight savings can still be achieved by having some of the structural components fabricated out of carbon fiber (e.g., the base plate), which would, nonetheless, increase the costs of the solution. 
Our membrane casting technique allows the creation of internal and external features on the membrane, which could potentially have interesting effects on the behavior of the gripper.
The design and the potential effects of those features could be a topic of future work.

As shown by our experiments, the presented gripper benefits from controlling the activation force during the grasping interval. 
Upcoming work thus features the development of a force controller that enables the tracking of the nominal activation force during the time the gripper is closing which is key for reliable grasping. 
Moreover, trajectory optimization will be used to guide the UAV to the payload under consideration of the particularities of the UG.

\section*{Appendix}
\label{sec:appendix}
\begin{figure}[t]
    \centering
    \resizebox{0.6\linewidth}{!}{%
    \begin{tikzpicture}
        \tikzset{
        ->, 
        node distance=2cm, 
        every state/.style={thick, fill=gray!10}, 
        initial text=power-up, 
        }
        \node[state, initial] (0) {undef.};
        \node[state, right of=0] (1) {startup};
        \node[state, right of=1] (2) {opened};
        \node[state, right of=2] (3) {closed};
        \node[state, above=of $(2.south)!0.5!(3.south)$] (4) {closing};
        \node[state, below=of $(2.north)!0.5!(3.north)$] (5) {opening};
        \draw
        (0) edge[left=0.3] node[below]{\MARKERCIRCLE{1}} (1)
        (1) edge[left=0.3] node[below]{\MARKERCIRCLE{2}} (2)
        (4) edge[above, bend left, left=0.5] node[right]{\MARKERCIRCLE{3}} (3)
        (2) edge[above, bend left, left=0.5] node[left]{\MARKERCIRCLE{4}} (4)
        (3) edge[above, bend left, left=0.5] node[right]{\MARKERCIRCLE{5}} (5)
        (5) edge[below, bend left, left=0.5] node[left]{\MARKERCIRCLE{6}} (2)
        (4) edge[in=105, out=75, loop] node[left]{} (4)
        (5) edge[in=105, out=75, loop] node[left]{} (5)
        (1) edge[in=105, out=75, loop] node[left]{} (1);
    \end{tikzpicture}}
    \caption{Gripper automaton. The gripper's behavior is governed by the state machine, transitioning between states once certain conditions are met. The main states are 'opened' and 'closed' with intermediary states to handle transitions in between. In the beginning, the state of the gripper is not known; therefore, it requires an initial boot procedure. Afterward, the state is well-defined by the data the sensors are providing.}
    \label{fig:automaton}
\end{figure}
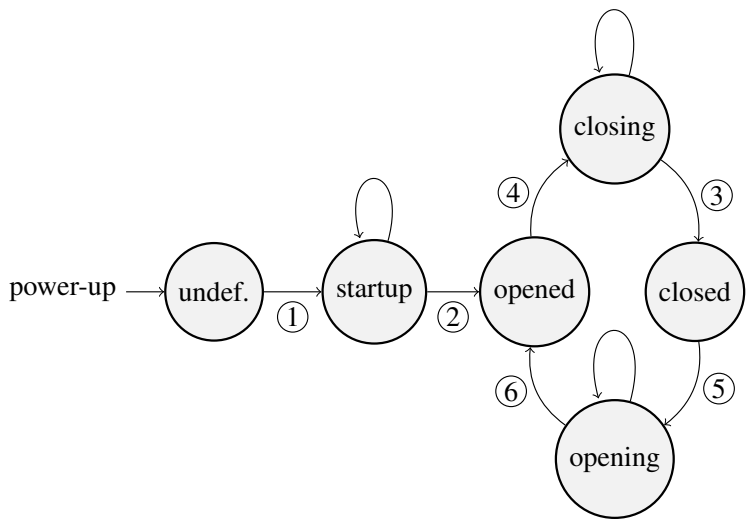

Herein the state machine (automaton) as it is depicted in \cref{fig:automaton} is described. 
At the beginning (power-up), we assume the state of the gripper to be undefined as it could be either jammed or fluidized. 
Thus, we start \MARKERCIRCLE{1} by running a 'startup' cycle which pulls all the air out of the membrane, then refills it until $P > P_{max}$ is reached, then enters the 'opened' state via \MARKERCIRCLE{2} indicating that the gripper is ready for operation. 
The transitions \MARKERCIRCLE{3} and \MARKERCIRCLE{6} depend on time and internal pressure. 
For the closing operation \MARKERCIRCLE{3} the condition is $P < P_{min}$ or $t > t_{vacc}$. 
Transition \MARKERCIRCLE{4} is automatically triggered if $F_m > F_{thr}$ is measured. 
The opening condition for transition \MARKERCIRCLE{6} is defined as $P > P_{max}$ or $t > t_{infl}$. 
The time condition is an additional safety feature in case of sensor malfunction.
State transition \MARKERCIRCLE{5} is typically triggered via user command over the serial port.
During the 'opening' and 'closing' states, one of the two pumps is running, pumping air either in or out of the system.

\section*{Funding}
This project was partially supported by the European Commission Horizon2020 research and innovation programme, under the grant agreement No 101017258 (SESAME) and by the Luxembourg National Research Fund (FNR) 5G-SKY project (ref. C19/IS/113713801).


\bibliographystyle{abbrv}
\bibliography{mybib}

\newpage


\end{document}